\documentclass[11pt]{article}
\usepackage[preprint]{acl}
\usepackage{microtype}
\usepackage{fontspec}
\newfontfamily\hangulfont{UnDotum.ttf}
\newfontfamily\hanjafont{UnDotum.ttf}
\usepackage{graphicx}
\usepackage{dblfloatfix}
\usepackage{cuted}
\usepackage{tabularx}
\usepackage{booktabs}
\usepackage{placeins}
\usepackage{enumitem}
\setlist[itemize]{topsep=1pt,partopsep=0pt}
\renewcommand{\textfraction}{0.05}
\newcommand{\fsamark}[3]{}
\providecommand{\tightlist}{\setlength{\itemsep}{0pt}\setlength{\parskip}{0pt}}
\title{SeenRef: How Reference Choice Distorts Few-Shot Style-Adaptation Scores}
\author{Jiwoo Jung \quad Daecheol An \\
  Sigmine \\
  \texttt{\{jiwoochris, andaniel\}@sigmine.ai}}
\begin{document}
\maketitle
\begin{abstract}
Few-shot style adaptation is evaluated by asking an authorship verifier how close a system's output is to reference texts by the target author, and the score is read as closeness to that author. We show that the score depends on which texts serve as the reference: when they are the system's own examples, as in the three protocols we read, it is not a reliable measure of closeness to the author. On a Korean and English benchmark where the author's own text is known, a reference that the adapter was trained on places its output closer to the author than unseen references do, and on LUAR above the author. Memorisation alone does not explain this: an adapter that reproduces its training texts in 1.7\% of cases has the same estimated inflation as one that does so in 95.5\%, and the inflation persists, somewhat reduced, when shared spans are masked. Two simple remedies fall short. Early stopping makes the inflation of the default adapter undetectable but degrades its outputs, and scoring against the author's own text with the same verifier confounds style with shared wording. We introduce SeenRef, which rescores outputs against several unseen texts by the author and reports the difference. With two simple filters, the same score selects outputs closer to the author's text. We detect no inflation with a larger training set or in a scaled-down rerun of AuthorMix.
\end{abstract}
\section{Introduction}

\begin{figure*}[t]\centering\includegraphics[width=\textwidth]{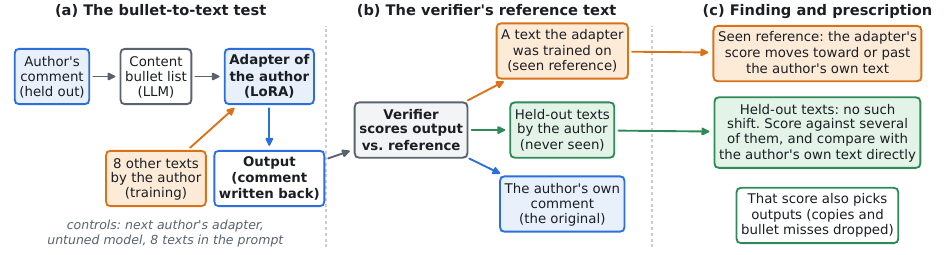}\caption{Overview. An adapter trained on an author's texts writes the author's held-out comment back from its content bullets (left), and a verifier scores the output against a reference text (centre). A reference the adapter was trained on inflates the score, whereas held-out texts and the author's own comment do not (right).}\end{figure*}

Few-shot style adaptation trains or prompts a model on a few texts by one person so that it writes new content in that person's style. The person's own text for the new content is rarely available, so the output is usually scored by an authorship verifier or a style embedding against reference texts by the person \citep{patel-2022-styll, horvitz-2024-tinystyler, thillainathan-2026-authormix}. The score is then read as a measure of how closely the system matches the author. We ask whether it can be read that way.

The answer depends on a choice that has received little attention: which texts serve as the reference. In all three studies, as far as the papers specify, the references are the examples the system was conditioned on. Two of the studies hold out the scoring model, but none holds out the references (Section 2). A held-out scorer is not a held-out reference. A trained adapter can reproduce parts of its training texts, and a verifier whose reference is one of those texts then scores this similarity together with the style. We do not rescore the published systems. Instead we isolate the effect of the reference in a controlled setting and rerun a scaled-down version of AuthorMix, the one system among the three that trains on its references.

Our setting is a benchmark where the author's own text for the same content is known. A language model writes a list of content bullets from each of 294 unedited Korean Wikipedia talk comments by 294 authors, and a LoRA adapter trained on up to eight other comments by the same author, which we call the own adapter, reconstructs the comment from its bullets. We compare it with an adapter trained on a different author and with the untuned model, and we replicate the experiment on 160 English authors. Each output is scored against four kinds of reference: the author's original comment, a text the adapter was trained on, one held-out text by the author, and the mean over four held-out texts. SeenRef, the procedure we propose, rescores outputs against held-out texts and reports the difference from the seen reference (Figure 2).

The choice of reference changes the conclusion. Scored against a training text, the own adapter is closer to the author than it is against held-out texts, and on English LUAR it scores above the author, whereas held-out references do not separate it from the author. Systems that were not trained on the reference gain less than half as much. The inflation of the adapter's score appears on our Korean verifier, on LUAR, on LambdaG and on prompted Claude Opus 5.5.

This is not simply memorisation. An adapter that reproduces its training texts in 1.7\% of cases has the same estimated inflation as one that reproduces them in 95.5\%, test outputs almost never share a long verbatim span with the reference, and the inflation persists, somewhat reduced, when the spans that they do share are masked. The two remedies that suggest themselves also fall short. Stopping the default adapter early on held-out loss makes the inflation undetectable but degrades its outputs, and scoring against the author's own text does not separate the author from the adapter on our verifier, while raw character overlap mostly measures the content that the bullets already carry. The inflation has limits: we do not detect it with 32 training texts or in the scaled-down AuthorMix rerun.

Our contributions are as follows.

\begin{itemize}
\tightlist
\item
  We show that the choice of reference texts distorts few-shot style scores. A reference the adapter was trained on brings the own adapter to 0.32 below the author on our Korean verifier, against 0.70 below with held-out references, and memorisation or copying alone does not account for the difference (Sections 4.1 and 4.2).
\item
  We show that two simple remedies, early stopping and scoring against the author's own text, do not solve the problem on their own (Section 4.3), and we propose SeenRef, an evaluation procedure that scores against several held-out texts by the author, reports the difference from the seen reference, and compares the output directly with the author's own text. Combined with two filters, the held-out score also selects outputs that are closer to that text than the default output (Section 4.4).
\item
  We build a topic-controlled Korean authorship benchmark, a bullet-to-text test in which the author's own text is the gold standard, and a Korean verifier for both. We will release the identifiers, prompts and scripts that rebuild the benchmark and retrain the verifier, but no model weights (Section 3).
\end{itemize}

\section{Related work}

\textbf{Scoring against the examples.} Few-shot authorship style transfer conditions a model on a few texts by an author and scores the output against them in an authorship embedding \citep{patel-2022-styll, khan-2023-stylemc, liu-2024-astrapop, liu-2026-grpo-authorial, horvitz-2024-tinystyler, thillainathan-2026-authormix, shin-2026-hyperstyler}. In the three protocols we read, the references are the texts the system was given. STYLL transfers ``using the target author's 16 posts as examples'' and scores movement toward an embedding of the same posts, on which copying one post scores the maximum. TinyStyler conditions on ``all 16 examples for the target author'', and AuthorMix fits its adapter mixture to its 16 reference sentences. Both score against those texts with a held-out embedding model. A held-out scorer is not a held-out reference, and we measure what the reference adds to trained adapters (Section 4).

\textbf{Scores that credit what the model was given.} In document-level LoRA, open-book scoring ``can credit a document's own adapter (its specialist) for reading that document as readily as for holding it'' \citep{pulici-2026-mostly-reading}, and an adapter trained on another document receives most of that credit. In our setting the credit goes to the adapter trained on the reference texts. In both cases the remedy is to score against what the system was not given.

\textbf{Korean and other agglutinative languages.} The corpus of \citet{cho-2022-stylekqc} covers style paraphrase rather than authorship. Japanese \citep{kanda-2025-japanese-ensemble, ghatpande-2026-japanese-lr, matsubara-2026-japanese-reviews} and Turkish \citep{akdag-2026-directed-attention} verification combines stylometric features with encoders or examples, and Korean work detects machine text \citep{park-2025-katfishnet, park-2026-korean-poetry}. To our knowledge no topic-controlled Korean authorship benchmark existed, so we build one.

\textbf{Authorship representations and verification.} Style embeddings \citep{rivera-soto-2021-luar, patel-2023-lisa, patel-2025-styledistance, qiu-2025-mstyledistance, kim-2025-multilingual-authorship, huang-2026-authbench}, which \citet{wang-2023-authorship-stylistic} examine for style versus topic, and content-independent tests \citep{wegmann-nguyen-2021-stel, wegmann-2022-same-author, kim-jurgens-2026-style-eliciting} stay near chance on our Korean pairs, and the seven languages of the benchmark of \citet{riverasoto-2026-steb} exclude Korean. Our public-code baselines are LambdaG \citep{nini-2026-lambdag}, which uses impostors as in \citet{koppel-winter-2014-impostors}, and linguistically informed prompting \citep{huang-etal-2024-llm-authorship}. Other work fine-tunes LLM verifiers \citep{kiefer-2026-avshift, kiefer-2026-gerav} or uses log-probabilities \citep{miralles-2025-oneshot-logprob}, and PAN scores verification by ROC-AUC, c@1, Brier and F-scores \citep{kestemont-2021-pan}. Obfuscation rewrites text against verifiers \citep{potthast-2016-author-obfuscation, bevendorff-2019-heuristic, mahmood-2019-mutantx, fisher-2024-styleremix}. Like \citet{zhu-jurgens-2021-idiosyncratic}, we perturb inputs to locate style (Appendix C).

\textbf{Style removal and imitation.} Style transfer often removes style first, by back-translation \citep{prabhumoye-etal-2018-style} or a neutral paraphrase \citep{krishna-etal-2020-reformulating}, as STYLL does for authors, and \citet{jin-etal-2022-deep} survey the field. Our content bullets play that role. In-context imitation is weak on informal text \citep{wang-etal-2025-catch}, and prompted impersonation evades neither LambdaG nor LUAR \citep{zeng-nini-2026-impersonation}.

\textbf{Gold-based evaluation.} References exist where a parallel corpus does, as for formality \citep{rao-tetreault-2018-dear}, where \citet{briakou-etal-2021-evaluating} validate metrics against human judgments, and LaMP \citep{salemi-etal-2024-lamp} scores personalised generation against the user's own text, whereas \citet{mir-2019-evaluating} scores intensity, content and naturalness without one. Our gold standard is the same author's held-out text for the same content, so a verifier score and a gold-based score can be read on one output.

\section{Experimental setup}

\subsection{Data}

The Korean data are talk-page comments from the archive of the Korean Wikipedia community forum ({\hangulfont 사랑방}). We define a work as all comments by one user on one archived discussion page and keep works of at least 250 characters after cleaning. The forum split has 8,299 works by 1,001 authors, divided into five author-disjoint folds. To test verifiers we build triples: an anchor and a positive text by one evaluation author, taken from different pages, and a negative text by another evaluation author, taken from the positive's page when one exists. The five folds hold 1,010 triples, 876 of them with a same-page negative, and each fold's verifier trains on 3,000 pairs from training authors only.

A second split applies the same rules to the 371 authors with at least 20,000 characters across the forum and four other talk namespaces. It has 40,696 works and 736 triples (Appendix B).

We chose this register because surface style carries a strong authorship signal in it. In a check that preceded the adapter experiments, character n-grams separate same-author from different-author pairs better than a content-only score by +0.073 AUC on these comments. For comparison, the same margin is +0.002 on the Blog Authorship Corpus \citep{schler-2006-blog} and between −0.022 and +0.060 on the Enron, blog, CCAT50 and Reddit sets of \citet{wang-etal-2025-catch}, so surface style matters more for authorship in our register than in these corpora.

\subsection{The bullet-to-text benchmark}

The benchmark reuses the comments, the 371 authors and the folds of the second split, and no trained component saw a fold's evaluation authors. We select 300 works, one per author, of 251 to 992 characters, and exclude 6, which leaves 294 works in five folds. For each work we keep a gold version of the original in which only spelling, spacing and typos are corrected (GPT-4.1). Claude Sonnet 5.5 writes the content bullets from the original, and the bullets retain 30.6\% of the original's words.

For each author we train a LoRA adapter on 4-bit Qwen3-8B \citep{yang-2025-qwen3} on up to eight of the author's other texts, each paired with its own bullets. We write k for the number of training texts, so k = 8 here.

Training stops after the first epoch whose mean training loss falls below 0.1, which all 294 adapters reach in 4 to 7 epochs. We call this the training-loss stop, and the adapter trained in this way the default adapter. Section 4 also uses a held-out stop, the epoch with the lowest loss on two held-out texts by the author. All systems decode greedily. We call an output degenerate if it reaches the token limit, is empty, or consists of more than 30\% repeated 20-character spans.

Four systems write each work (Figure 1): the own adapter, the next-author adapter (the adapter of the next author in the fold, which serves as a mismatched control), the untuned model, and the untuned model prompted with eight of the author's texts. Appendix B shows one work with its bullets and outputs.

\subsection{Reference texts and the inflation index}

We score each output against four kinds of reference text. The first is the original itself, in its gold version. The second is a training reference, one text the own adapter was trained on, as in the protocols of Section 2. The third is a held-out reference, another text by the author that no system saw. The fourth is the mean score over four held-out references.

The inflation index of a system x compares its distance from the original under the two kinds of reference:

\[\mathrm{II}(x) = [s_t(x) - s_t(\mathrm{orig})] - [s_h(x) - s_h(\mathrm{orig})],\]

where \(s_t\) is the score against the training reference and \(s_h\) the score against the held-out references. Subtracting the original's score removes the shift that every text receives when the reference changes, so II retains only what the training reference adds for system x. For example, on our Korean verifier the own adapter is −0.32 from the original against the training reference and −0.70 against the held-out mean, which gives II = +0.38.

\subsection{Verifiers and statistics}

Our Korean verifier, which we call M2, is a LoRA pair classifier on Qwen3-8B, trained per fold on same-author and different-author pairs. It scores two 460-token texts by log p(yes)~− log p(no), averaged over both orders. On the 1,010 topic-matched triples by unseen authors it reaches a pooled ROC-AUC of 0.8263. This is +0.0990 above the best of twelve baselines, character n-gram TF-IDF at 0.7273, while public encoders stay near chance; prompted Claude Opus 5.5 scores higher (Appendix C). We use M2 because it runs locally on an open model and was fixed before the adapter experiments. Its scores are uncalibrated and their scale varies by fold, so we interpret M2 differences as signs and orderings rather than magnitudes.

In English we use LUAR \citep{rivera-soto-2021-luar}, Style-Embedding \citep{wegmann-2022-same-author} and EN-M2, the same recipe trained on English pairs. LambdaG \citep{nini-2026-lambdag} scores a text against a set of known texts by the author, which plays the role of the reference. We also report chrF \citep{popovic-2015-chrf} and the cosine of word-ending frequencies between the output and the original.

Unless stated otherwise, we report differences as the own adapter minus the other system, with 95\% paired bootstrap intervals. For 32 comparisons we wrote down the decision rule and committed it to our private repository before the corresponding run, and timestamp proofs date the repository on \mbox{2026-10-25} and \mbox{2026-11-13}. Appendix A lists these pre-specified comparisons with their outcomes. They concern the verifier and the bullet-to-text test, and several belong to experiments whose results appear only in the ancillary ledger.

The reference-text comparisons of Section 4 also followed rules fixed before scoring, and we report them as estimates with intervals rather than as tests. Other analyses were not planned in advance. The appendix reports the supporting results. The ancillary files of this submission hold the complete record of every check, which we cite as ledger A.1 to A.18, and a supplement with the commands behind each number, which we cite as S1.

\begin{table*}[!t]\centering\small\setlength{\tabcolsep}{3.5pt}\begin{tabular}{llrrrrr}\toprule
Verifier & Reference & n & Own~− original & Own~− next & Own~− untuned & II (own) \\\midrule
Korean M2 & original (gold) & 294 & \begin{tabular}[t]{@{}r@{}}+0.031\\{\scriptsize {[}−0.036, +0.094{]}}\end{tabular} & \begin{tabular}[t]{@{}r@{}}+0.290\\{\scriptsize {[}+0.208, +0.377{]}}\end{tabular} & \begin{tabular}[t]{@{}r@{}}+0.376\\{\scriptsize {[}+0.308, +0.441{]}}\end{tabular} &  \\
Korean M2 & 1 training text & 266 & \begin{tabular}[t]{@{}r@{}}−0.321\\{\scriptsize {[}−0.479, −0.166{]}}\end{tabular} & \begin{tabular}[t]{@{}r@{}}+0.898\\{\scriptsize {[}+0.742, +1.069{]}}\end{tabular} & \begin{tabular}[t]{@{}r@{}}+1.523\\{\scriptsize {[}+1.375, +1.668{]}}\end{tabular} &  \\
Korean M2 & 4 held-out (mean) & 266 & \begin{tabular}[t]{@{}r@{}}−0.704\\{\scriptsize {[}−0.819, −0.592{]}}\end{tabular} & \begin{tabular}[t]{@{}r@{}}+0.623\\{\scriptsize {[}+0.503, +0.750{]}}\end{tabular} & \begin{tabular}[t]{@{}r@{}}+1.175\\{\scriptsize {[}+1.069, +1.285{]}}\end{tabular} & \begin{tabular}[t]{@{}r@{}}+0.384\\{\scriptsize {[}+0.249, +0.526{]}}\end{tabular} \\
\addlinespace[3pt]
EN-M2 & 1 training text & 960 (160) & \begin{tabular}[t]{@{}r@{}}−0.146\\{\scriptsize {[}−0.238, −0.055{]}}\end{tabular} & \begin{tabular}[t]{@{}r@{}}+0.779\\{\scriptsize {[}+0.673, +0.894{]}}\end{tabular} & \begin{tabular}[t]{@{}r@{}}+1.051\\{\scriptsize {[}+0.949, +1.161{]}}\end{tabular} &  \\
EN-M2 & 1 held-out text & 960 (160) & \begin{tabular}[t]{@{}r@{}}−0.234\\{\scriptsize {[}−0.327, −0.141{]}}\end{tabular} & \begin{tabular}[t]{@{}r@{}}+0.610\\{\scriptsize {[}+0.499, +0.729{]}}\end{tabular} & \begin{tabular}[t]{@{}r@{}}+0.871\\{\scriptsize {[}+0.758, +0.990{]}}\end{tabular} & \begin{tabular}[t]{@{}r@{}}+0.087\\{\scriptsize {[}+0.008, +0.168{]}}\end{tabular} \\
\addlinespace[3pt]
English LUAR & 8 training texts & 960 (160) & \begin{tabular}[t]{@{}r@{}}+0.0223\\{\scriptsize {[}+0.0132, +0.0315{]}}\end{tabular} & \begin{tabular}[t]{@{}r@{}}+0.0542\\{\scriptsize {[}+0.0429, +0.0662{]}}\end{tabular} & \begin{tabular}[t]{@{}r@{}}+0.0344\\{\scriptsize {[}+0.0247, +0.0450{]}}\end{tabular} &  \\
English LUAR & 8 held-out texts & 960 (160) & \begin{tabular}[t]{@{}r@{}}+0.0035\\{\scriptsize {[}−0.0053, +0.0122{]}}\end{tabular} & \begin{tabular}[t]{@{}r@{}}+0.0321\\{\scriptsize {[}+0.0205, +0.0441{]}}\end{tabular} & \begin{tabular}[t]{@{}r@{}}+0.0147\\{\scriptsize {[}+0.0046, +0.0243{]}}\end{tabular} & \begin{tabular}[t]{@{}r@{}}+0.0189\\{\scriptsize {[}+0.0130, +0.0251{]}}\end{tabular} \\
\addlinespace[3pt]
Korean Opus 5.5 & 1 held-out text & 267 &  & \begin{tabular}[t]{@{}r@{}}+0.088\\{\scriptsize {[}+0.063, +0.112{]}}\end{tabular} & n/a & \begin{tabular}[t]{@{}r@{}}+0.104\\{\scriptsize {[}+0.076, +0.132{]}}\end{tabular} \\
\addlinespace[3pt]
Korean LambdaG & 8 training (known) & 217 & \begin{tabular}[t]{@{}r@{}}+6.10\\{\scriptsize {[}+4.07, +7.94{]}}\end{tabular} & n/a & n/a &  \\
Korean LambdaG & 8 held-out (known) & 217 & \begin{tabular}[t]{@{}r@{}}−3.13\\{\scriptsize {[}−4.84, −1.56{]}}\end{tabular} & n/a & n/a &  \\
\addlinespace[3pt]
chrF & none (direct) & 294 & n/a & \begin{tabular}[t]{@{}r@{}}+1.44\\{\scriptsize {[}+0.22, +2.64{]}}\end{tabular} & \begin{tabular}[t]{@{}r@{}}−1.52\\{\scriptsize {[}−2.46, −0.67{]}}\end{tabular} &  \\
\addlinespace[3pt]
Bullet-masked chrF & none (direct) & 294 & n/a & \begin{tabular}[t]{@{}r@{}}+1.26\\{\scriptsize {[}+0.53, +2.03{]}}\end{tabular} & \begin{tabular}[t]{@{}r@{}}+2.43\\{\scriptsize {[}+1.75, +3.10{]}}\end{tabular} &  \\\bottomrule
\end{tabular}\caption{The own adapter minus the author's original, the next-author adapter and the untuned model, under different reference texts (mean {[}95\% interval{]}). II is the inflation index of the own adapter (Section 3.3). n is the number of works, with authors in parentheses. For the Opus row the difference from the original was computed in the opposite direction: the original scores +0.226 above the own adapter.}\end{table*}
\begin{table*}[!t]\centering\small\setlength{\tabcolsep}{3.5pt}\begin{tabular}{llrrr}\toprule
Difference & Score & Qwen3-8B & Gemma 3 12B & Qwen3-32B \\\midrule
own~− next author's adapter & M2 score & +0.290 & +0.339 & +0.233 \\
own~− untuned model & M2 score & \begin{tabular}[t]{@{}r@{}}+0.376\\{\scriptsize {[}+0.308, +0.441{]}}\end{tabular} & \begin{tabular}[t]{@{}r@{}}+0.551\\{\scriptsize {[}+0.477, +0.627{]}}\end{tabular} & \begin{tabular}[t]{@{}r@{}}+0.653\\{\scriptsize {[}+0.579, +0.727{]}}\end{tabular} \\
own~− 8 texts in the prompt & M2 score & +0.798 & +0.434 & +0.610 \\
own~− original & M2 score & \begin{tabular}[t]{@{}r@{}}+0.031\\{\scriptsize {[}−0.036, +0.094{]}}\end{tabular} & \begin{tabular}[t]{@{}r@{}}−0.047\\{\scriptsize {[}−0.127, +0.028{]}}\end{tabular} & \begin{tabular}[t]{@{}r@{}}−0.040\\{\scriptsize {[}−0.108, +0.028{]}}\end{tabular} \\
own~− next author's adapter & M2, held-out reference & \begin{tabular}[t]{@{}r@{}}+0.600\\{\scriptsize {[}+0.456, +0.746{]}}\end{tabular} & n/a & n/a \\
own~− original & M2, held-out reference & \begin{tabular}[t]{@{}r@{}}−0.829\\{\scriptsize {[}−0.962, −0.694{]}}\end{tabular} & n/a & n/a \\
\addlinespace[3pt]
own~− untuned model & chrF to the original & \begin{tabular}[t]{@{}r@{}}−1.52\\{\scriptsize {[}−2.46, −0.67{]}}\end{tabular} & n/a & n/a \\
own~− untuned model & ending cosine to the original & \begin{tabular}[t]{@{}r@{}}−0.021\\{\scriptsize {[}−0.038, −0.006{]}}\end{tabular} & n/a & n/a \\
own~− original & LambdaG, training texts known & \begin{tabular}[t]{@{}r@{}}+7.37\\{\scriptsize {[}+5.54, +9.11{]}}\end{tabular} & \begin{tabular}[t]{@{}r@{}}+9.54\\{\scriptsize {[}+7.88, +11.34{]}}\end{tabular} & \begin{tabular}[t]{@{}r@{}}+7.78\\{\scriptsize {[}+4.93, +10.25{]}}\end{tabular} \\
own~− original & LambdaG, held-out texts known & \begin{tabular}[t]{@{}r@{}}−3.13\\{\scriptsize {[}−4.84, −1.56{]}}\end{tabular} & n/a & n/a \\\bottomrule
\end{tabular}\caption{Bullet-to-text on 294 works: the own adapter (k = 8) minus each other system and the original, with 95\% intervals for the main contrasts. M2 is the Qwen3-8B verifier for every base model. Held-out rows use one held-out reference on 267 works, whereas Table 1 uses the mean over four. The LambdaG training-text row uses 270, 260 and 273 works for the three base models, so its first value differs from Table 1, which uses the 217 works that also have eight held-out texts. n/a: not computed.}\end{table*}
\section{How the reference changes the score}

\begin{figure*}[t]\centering\includegraphics[width=\textwidth]{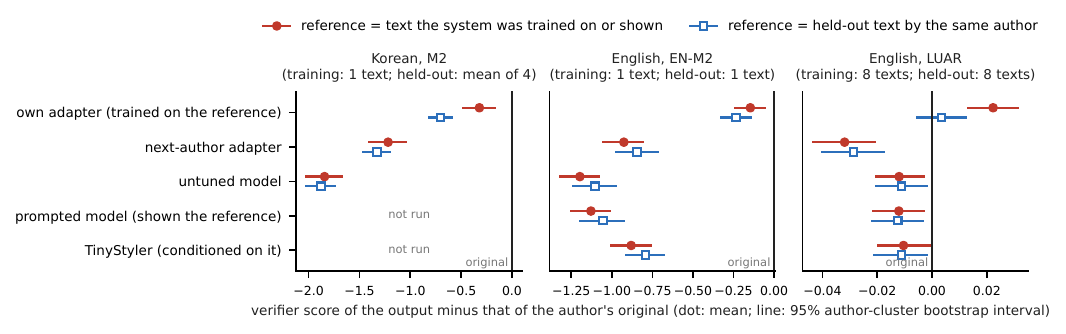}\caption{Output score minus the score of the author's own comment, per system, when the verifier's reference is a text the adapter was trained on (filled) or a held-out text (open). Bars are 95\% author-cluster intervals. The adapter trained on the reference gains most.}\end{figure*}

\subsection{Seen references inflate the score}

Table 1 and Figure 2 show the same outputs under different references. With a training text as the reference, the Korean verifier places the own adapter 0.32 below the author's original. With the mean over four held-out references, it places the same outputs 0.70 below. On 160 English authors the pattern repeats on EN-M2, and on LUAR the own adapter scores above the author with training references and level with the author with held-out ones. The next-author adapter and the untuned model gain less than half as much as the own adapter (Figure 2), and on EN-M2 their scores fall slightly, with intervals that include zero. Systems that are shown the references without being trained on them, namely the prompted model and TinyStyler, show no detectable inflation on LUAR or EN-M2. On Style-Embedding, the embedding that TinyStyler conditions on, the pattern differs: TinyStyler's inflation index is above zero, whereas the interval of the own adapter's index includes zero (Appendix F). The inflation also appears when prompted Claude Opus 5.5, the strongest verifier in our comparison, replaces M2 (Table 1), although Opus may have seen these comments in pre-training.

Other verifiers show the same pattern. LambdaG scores the own adapter above the author when its known texts are the adapter's training texts, and below the author when they are eight held-out texts (Table 1, same 217 works), although we did not test the cause of this difference for LambdaG. LLM judges pick the own adapter over the next-author adapter in 75.4\% of judgments when shown training texts and in 63.4\% when shown held-out texts (147 items; Appendix E). The inflation on M2 is also stable across two further seeds and two further base models, where II lies between +0.31 and +0.45.

\subsection{Memorisation alone does not explain the inflation}

The obvious explanation is that the adapter has memorised the reference. Figure 3 tests it. It gives the inflation index under different training settings, next to the share of outputs that reproduce a 100-character span of a training text when the adapter is given its own training bullets. At the training-loss stop the adapter reproduces its training texts verbatim in 95.5\% of cases and II is clearly positive. Memorisation is nevertheless neither necessary nor sufficient. An adapter trained for two fixed epochs reproduces its training texts in only 1.7\% of cases and has the same estimated II as the adapter at the training-loss stop (Figure 3). Two further settings that rarely memorise (rank 2, and a lower learning rate at rank 16, which also stops on held-out loss) keep the interval of II above zero as well, whereas the adapter trained on 32 texts memorises in 96.2\% of cases and the interval of its II includes zero.

\begin{figure}[!tp]\centering\includegraphics[width=\columnwidth]{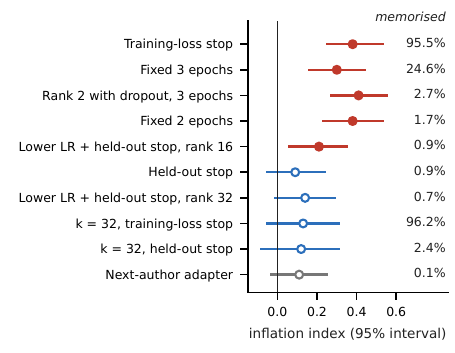}\caption{Inflation index on Korean M2 by training setting of the own adapter, with 95\% intervals (filled: interval above zero); the last row is the next-author adapter. Appendix D gives the values and sample sizes. Memorisation is the share of outputs that repeat a 100-character span of a training text when the adapter is given its training bullets. Settings use eight training texts unless noted, and LR is the learning rate.}\end{figure}

Copying at test time is not the explanation either. Only one test output shares a 40-character span with its training reference. When we mask the character 8-grams that each text shares with its reference, the own adapter's II remains above zero. With one held-out reference it is +0.43 {[}+0.28, +0.58{]} after masking, against +0.52 before (Appendix E). The topic of the reference does not explain the inflation either. We did not test whether output length or the quality of fit drives both quantities.

What the inflation does follow, weakly, is similarity. Across the 266 Korean works of Table 1, we compute the own adapter's II for each work and correlate it with a similarity measure, namely the output's chrF to the training reference minus the original's chrF to the same reference. Spearman's ρ is +0.199 {[}+0.079, +0.313{]}, and the mean II rises from 0.19 in the lowest quartile of similarity to 0.68 in the highest.

\subsection{Two simple remedies fall short}

\textbf{Early stopping.} If the inflation came from overfitting, stopping earlier would be the remedy. At eight training texts the held-out stop does remove the memorisation, and the inflation is no longer detectable (+0.09, with an interval that includes zero). It has a cost, however: its margin over the next-author adapter is no longer separated from zero when the original is the reference, 32.6\% of its outputs are degenerate, against 6.8\% at the training-loss stop, and its outputs are farther from the original than those of the untuned model (Appendix D). With a lower learning rate, stopping on held-out loss reduces the inflation without removing it: at rank 16 the interval of II stays above zero, and at rank 32 it includes zero (Figure 3).

\textbf{The original as the reference.} Where the author's own text exists, scoring against it seems the natural remedy, but it is not sufficient on its own. With the original itself as the reference, M2 does not rank the original above the adapter's output (own~− original +0.031, with an interval that includes zero; Table 1), although the adapter never saw the original. In this setting M2 mainly rewards shared wording: degraded copies of the original score lower the fewer of its words they keep (Appendix E). On raw chrF the own adapter is farther from the original than the untuned model (−1.52, with an interval that excludes zero; Table 1). This is largely an effect of the bullets. Scored as if it were an output, the bullet list alone reaches a chrF of 27.91, against 27.87 for the own adapter, so raw chrF mostly measures content that the bullets already carry. When we drop the bullet words from both the output and the original, the own adapter is closer to the original than the untuned model (+2.43) and the next-author adapter (+1.26). We call this measure bullet-masked chrF.

\textbf{Single references.} One held-out reference is not enough either. It places the adapter lower than the mean over four does: the own adapter is then 0.83 below the original, and on a later sample of 174 new authors the difference has the same sign, but its interval includes zero (Appendix D). We therefore score against the mean over several held-out references.

\subsection{SeenRef as a selection rule}

The held-out score also selects among candidate outputs without reading the original. For each author we choose among seven adapters trained with different stopping rules and budgets. We drop outputs that miss their content bullets or copy a training text and then take the output with the highest M2 against the held-out mean. On bullet-masked chrF, this choice beats the greedy output of the adapter trained with the training-loss stop by +0.88 {[}+0.29, +1.47{]}, and on raw chrF by +1.58. The two filters matter: without them the highest-scoring output moves away from the original (raw chrF −2.68).

Applying the same rule to eight sampled outputs of one adapter also beats its greedy output, by +1.12 {[}+0.59, +1.66{]}. In 19 works no sample passes the filters and the greedy output is kept. Training adapters further on these choices did not help in a 50-work pilot (Appendix D).

\subsection{AuthorMix rerun}

We reran a scaled-down AuthorMix, the one system among the three protocols that optimises on its reference sentences. It does not gain detectably more than its base model from being scored against those 16 sentences: the difference in II is −0.0008 {[}−0.0025, +0.0009{]} over 10 target authors (Appendix D). Our rerun trains far less than the original system, so this result does not test the published configuration.

\section{What the adapter learns}

The inflation of Section 4 would matter little if the adapters did not capture their authors at all. Table 2 shows that they do, and also where they fall short.

With the original as the reference, M2 scores the own adapter +0.290 above the next-author adapter, and the own adapter is higher in 70.6\% of works. The margin survives when copied strings are removed: when we mask every character 8-gram that an output shares with its adapter's training texts, the own adapter stays above the next-author adapter by +0.205 {[}+0.125, +0.284{]}. The margin is larger, at +0.600, when the reference is a held-out text by the author, and it is +0.242 across three adapter seeds. It holds on Gemma 3 12B and Qwen3-32B and on English Wikipedia talk comments, where LUAR and EN-M2 both place the own adapter above the next-author adapter on 160 authors (Appendix F). A character n-gram verifier that uses no LLM agrees (+0.252).

Two results limit this picture. Topic may explain part of the margin. We look at the works for which the next author's training texts are at least as close in topic to the target as the own author's. With an embedding measure of topic there are 71 such works and the margin is +0.233. With a word-stem measure there are 38 such works, and the margin is +0.046 with an interval that includes zero. Prompting is also a strong alternative: GPT-6.1 Sol prompted with the same eight texts is not separated from the own adapter on M2 and is closer to the original on chrF, although it may have seen the comments in pre-training (Appendix E). These adapters were trained with the training-loss stop, and with the held-out stop the margin is not separated from zero. A further experiment in the ledger tempers the reading of Table 2: when author adapters are trained on top of an adapter shared across authors, the author step adds nothing on chrF over the shared adapter, so much of the gain over the untuned model reflects register rather than the individual author (ledger A.15). Results for other numbers of training texts, training budgets and verifier variants are in the ledger as well.

\section{Conclusion}

The texts that a verifier receives as the author's reference can determine what a few-shot style score measures, and for a system trained on its examples, a score obtained with those examples as references is not a reliable measure of closeness to the author. When the reference is a text the adapter was trained on, the adapter's score moves toward the author's own text, and on LUAR beyond it, while systems not trained on that text gain less than half as much. The effect is clearest for adapters that nearly memorise their training texts, and we do not detect it when the default adapter is stopped on held-out loss or in a scaled-down AuthorMix rerun. We recommend three practices. First, the texts a system was trained on should not serve as the verifier's reference. Second, outputs should be scored against the mean over several texts by the author that the system never saw, and the number of texts should be reported. Third, where the author's own text for the same content exists, the output should be compared with it directly, next to a mismatched adapter and the untuned model. The bullet-to-text benchmark supplies such texts.

\leavevmode\pdfsavepos\write\csname @auxout\endcsname{\string\fsamark{ZZBODYENDZZ}{\thepage}{\the\pdflastxpos}}

\section*{Limitations}
\addcontentsline{toc}{section}{Limitations}

The inflation is clearest for adapters trained until they nearly reproduce their training texts, although three settings that rarely memorise also keep it above zero (Figure 3). With a held-out stop at eight texts and with 32 training texts, the interval of the inflation index includes zero, and so does the interval for AuthorMix against its base model in our scaled-down rerun. We did not rerun the published evaluations of these systems: AuthorMix was rerun only in a scaled-down form and TinyStyler only on our English data, so our results do not show that published scores are inflated.

M2 is uncalibrated, uses one model size and truncates inputs to 460 tokens, and each fold has its own verifier. We therefore read its differences as signs, not sizes. M2 is below prompted Claude Opus 5.5 on the same pairs, and our contribution is a scoring procedure, not a better verifier. No human evaluation has been run. We designed a rating task but did not carry it out. LLM judges rated its items, but they do not replace human raters.

One model with one prompt writes the bullets, which retain 30.6\% of the original's words, so the test separates content from wording only in part. Tool definitions from an account connector entered the prompt of 14.6\% of the bullet lists, and the main results hold without those works.

Each work has a single control author. Both splits share the register of Korean Wikipedia, and the English evidence is limited to 200 Wikipedia talk users at k = 8. The base models may have seen these public comments during pre-training, since 287 of the 294 works predate their release. The own and next-author adapters share that exposure, but the comparison with the untuned model does not separate it from adaptation.

Apart from two settings with a lower learning rate, all adapters share one learning rate and batch size, and only the main Qwen3-8B adapters have three seeds. The dependence of the inflation on stopping rules and budgets may differ in other settings.

The planned comparisons are the 32 of Appendix A, whose rules were committed to a private repository rather than a public registry. Each is read on its own interval without correction across comparisons, although the 28 whose intervals exclude zero survive Holm and Benjamini--Hochberg corrections applied afterwards. The reference-text results of Section 4 are estimates and should be read as such. Appendix G gives further limitations.

\section*{Ethics statement}
\addcontentsline{toc}{section}{Ethics statement}

The benchmark uses public Korean and English Wikipedia discussion comments under CC BY-SA 4.0, and the diagnostic studies use free-licence Wikisource works and the Blog Authorship Corpus. We will release identifiers, hashes, prompts and rebuild scripts rather than texts, when the code is published. Author identifiers are hashes of user names and are pseudonymous, not anonymous. Wikipedia contributors cannot be screened by age, so the data may include minors.

Comment texts were sent to commercial LLM services for normalisation, bullet writing, baselines and judging (Appendix H lists the services). The adapters and M2 ran locally on one desktop machine.

A trained verifier can link texts written under different accounts, and a per-author adapter can impersonate a contributor. An adapter given its own training bullets returns the training text word for word in 46.4\% of cases. For these reasons we release no verifier weights, adapter weights or author-level outputs, and we report pooled results only. No authorship verifier or AI-text detector is used in training or as a reward, apart from one 50-work pilot that tuned adapters on outputs chosen by M2. The intended use of this work is to check how style adaptation is scored, not to impersonate people, to hide machine writing or to evade detection.

The paper quotes one public comment with its bullets and two outputs, credited by a revision URL in the ancillary file S1, as the Wikimedia Terms of Use allow. No human evaluation was run and no institutional ethics review was sought, since the study uses public texts and has no human participants.

\bibliographystyle{acl_natbib}

\appendix
\onecolumn
\raggedbottom
\setlength{\intextsep}{2pt plus 1pt minus 1pt}\setlength{\textfloatsep}{4pt plus 1pt minus 1pt}\setlength{\floatsep}{4pt plus 1pt minus 1pt}\renewcommand{\topfraction}{0.95}\renewcommand{\bottomfraction}{0.95}\renewcommand{\textfraction}{0.05}
\leavevmode\pdfsavepos\write\csname @auxout\endcsname{\string\fsamark{ZZAPPSTARTZZ}{\thepage}{\the\pdflastxpos}}

\setcounter{table}{0}\renewcommand{\thetable}{A\arabic{table}}
\FloatBarrier\section*{A\quad Pre-specified comparisons}
\addcontentsline{toc}{section}{Appendix A: Pre-specified comparisons}

We call a comparison pre-specified when its decision rule was committed to our private version-control repository before the corresponding run. We did not use a public registry, and ledger A.9 describes the procedure. There are 32 pre-specified comparisons, and they concern the verifier and the bullet-to-text test. The reference-text comparisons of Section 4, which carry the paper's main claim, are not among the 32, and we report them as estimates under rules fixed before scoring. In the tables of the later appendices a dagger (†) marks these planned comparisons. Three further comparisons were fixed in the same way but lie outside the 32, and their tables say so: two for the English expansion (Appendix F) and the endpoint of the AuthorMix rerun (Appendix D).

OpenTimestamps proofs of the repository heads show that the rules of 30 of the 32 comparisons existed by \mbox{2026-10-25} and all 32 by \mbox{2026-11-13}. The other two, those of the shared-adapter variant H1 and the word-end verifier H2, were committed on \mbox{2026-10-27} and \mbox{2026-10-28}. The proofs fix the commits as of \mbox{2026-10-25} and \mbox{2026-11-13}, the dates of the proofs, and not the time of each commit. For each rule whose commit the ancillary file S1 lists (31 commits), the arXiv ancillary files hold an English excerpt of the rule with its full commit hash and time. The excerpt is a copy of the rule text and not the git history, and the other rules have no excerpt there.

Table A1 lists the thirteen comparisons whose experiments are reported in this paper. Five concern the verifier (Section 3.4) and eight the bullet-to-text test (Sections 3.2 and 5). The surface-feature verifier M1 and M2 pass against the best baseline, but on the literary corpus M2 is below the word-final baseline. In the bullet-to-text test the own adapter is above the next-author adapter and above the prompted model on all three base models, and both English comparisons on LUAR are positive.

\begin{table}[!htb]\centering\small\setlength{\tabcolsep}{5pt}\begin{tabular}{lrrl}\toprule
Comparison & n & Difference {[}interval{]} & Outcome \\\midrule
\multicolumn{4}{l}{\emph{Verifier, ROC-AUC, 97.5\% intervals}} \\
M1~− best baseline & 1,010 & +0.0317 {[}+0.0075, +0.0550{]} & passes \\
M2~− best baseline, main split & 1,010 & +0.0990 {[}+0.0783, +0.1223{]} & passes \\
M2~− best baseline, seed 1 & 1,010 & +0.1068 {[}+0.0865, +0.1290{]} & passes \\
M2~− best baseline, second split & 736 & +0.0688 {[}+0.0385, +0.0988{]} & passes \\
M2~− word-final, literary & 358 & −0.0646 {[}−0.1095, −0.0184{]} & below \\
\addlinespace[2pt]
\multicolumn{4}{l}{\emph{Bullet-to-text, 95\% intervals}} \\
Own~− next, Qwen3-8B & 294 & +0.290 {[}+0.208, +0.377{]} & positive \\
Own~− next, Gemma 3 12B & 294 & +0.339 {[}+0.242, +0.438{]} & positive \\
Own~− next, Qwen3-32B & 294 & +0.233 {[}+0.155, +0.315{]} & positive \\
Own~− prompted, Qwen3-8B & 294 & +0.798 {[}+0.727, +0.867{]} & positive \\
Own~− prompted, Gemma 3 12B & 294 & +0.434 {[}+0.360, +0.508{]} & positive \\
Own~− prompted, Qwen3-32B & 294 & +0.610 {[}+0.538, +0.683{]} & positive \\
English, LUAR: own~− next & 40 & +0.0480 {[}+0.0352, +0.0612{]} & positive \\
English, LUAR: own~− prompted & 40 & +0.0194 {[}+0.0067, +0.0314{]} & positive \\\bottomrule
\end{tabular}\caption{The thirteen pre-specified comparisons reported in this paper. Verifier rows are ROC-AUC differences on n topic-matched triples. Bullet-to-text rows are M2 differences at k = 8 on n works, or LUAR differences on n English authors. Next: next-author adapter. Prompted: prompted model.}\end{table}

The other nineteen comparisons belong to experiments that are reported only in the ledger: the number of training texts, the stopping rules and training budgets, the imitation-aware verifier M2\ensuremath{'} (a verifier retrained on machine imitations), the word-end verifier H2 and the shared adapter (ledger A.14 and A.15). Their outcomes are not all favourable. The imitation-aware verifier and the word-end verifier fail their rules. The margin of the own adapter over the next-author adapter is not separated from zero at two training texts and under the held-out stop. Under the held-out stop, two fixed epochs and the rank-2 setting, the output is farther from the original than the untuned model's. Ledger A.17 lists all 32 comparisons with their outcomes. The rule excerpts are in the ancillary files, and the ancillary file S1 lists the commits and the result files.

\FloatBarrier\section*{B\quad Data and benchmark construction}
\addcontentsline{toc}{section}{Appendix B: Data and benchmark construction}

\subsection*{B.1 Corpus and splits}
\addcontentsline{toc}{subsection}{B.1 Corpus and splits}

The source is the archive of talk-page comments from the Korean Wikipedia community forum ({\hangulfont 사랑방}). A work is the concatenation of all comments by one user on one archived discussion page, cleaned of HTML escapes and signatures and kept if it has at least 250 characters. We split the authors into 5 folds, and the overlap between evaluation and training authors is 0 in every fold. For each evaluation author with several works we build up to 2 pairs of an anchor and a positive from different pages. The negative is a work by another evaluation author on the same page as the positive, or from the same year if none exists.

Each fold's verifier trains on pairs drawn from training authors only, and its validation pairs come from 40 held-out training authors. With so few authors, 30.4\% of the validation negatives come from the same year against 2.9\% of the training negatives, so validation loss measures an easier task than training loss. The second split applies the same rules to the authors whose works in the forum and in four other talk namespaces (Talk, User talk, Wikipedia and Wikipedia talk) total at least 20,000 characters. As Table A2 shows, it has fewer same-page negatives, because user-talk pages have few participants. Fold-level counts, page classes and signature patterns are in ledger A.7.

\begin{table}[!htb]\centering\footnotesize\setlength{\tabcolsep}{5pt}\begin{tabular}{lrr}\toprule
Quantity & Forum split & Second split \\\midrule
Authors & 1,001 & 371 \\
Works & 8,299 & 40,696 \\
Works over 460 tokens & 36.2\% & 37.8\% \\
Triples & 1,010 & 736 \\
Same-page negatives & 876 & 444 (60.3\%) \\
Negatives from any other author & 1 & 0 \\
Training pairs per fold & 3,000 & 3,000 \\
Validation pairs per fold & 200 & 200 \\
\addlinespace[2pt]
\multicolumn{3}{l}{\emph{Selection of the bullet-test works}} \\
Authors removed: used by earlier pilots &  & 56 \\
Authors removed: no eligible work or too few exemplar candidates &  & 15 \\
Works selected &  & 300 \\
Works excluded: exemplar overlap &  & 1 \\
Works excluded: normalisation above the 5\% threshold &  & 1 \\
Works excluded: dash signature in the original &  & 4 \\
Works evaluated &  & 294 \\\bottomrule
\end{tabular}\caption{The two splits after cleaning, and the selection of the bullet-test works from the authors of the second split. Triples and negatives are totals over five folds. The second split was fixed before its run (Appendix A).}\end{table}

\subsection*{B.2 Evaluation works and prompts}
\addcontentsline{toc}{subsection}{B.2 Evaluation works and prompts}

The bullet-to-text test (Sections 3.2 and 5) uses one work per author. We remove the authors used by earlier pilots and those that lack a work of 250 to 1,000 characters or enough exemplar candidates. The 300 works that remain are all selected, and 6 are kept out of evaluation for the reasons in the lower part of Table A2. The gold version is the original with only spelling, spacing and typos corrected by GPT-4.1, and this normalisation changes 0.0088 of the characters on average. The bullet prompt asks for the facts, claims and requests of a text as neutral noun phrases that carry over none of its tone, endings or wording. The writing prompt, the same for all systems, asks for a Wikipedia talk comment with only the listed content. The prompted model also receives 8 texts by the author. The verbatim prompts are in ledger A.7.

\subsection*{B.3 Example and evaluation sets}
\addcontentsline{toc}{subsection}{B.3 Example and evaluation sets}

Table A3 shows one work in English translation, with its bullets, the gold original, and the outputs of the own adapter and the next-author adapter. The Korean text and the contributor credit are in the ancillary file S1. The evaluation sets of the paper are listed in Table A4, and each difference that we report is paired within one set.

\begin{table}[!htb]\centering\small\setlength{\tabcolsep}{5pt}\begin{tabularx}{\textwidth}{l>{\raggedright\arraybackslash}X}\toprule
Text & English translation (ours) \\\midrule
Bullets & critical stance on one-sided, contrived translation practice and the claim that it should be avoided / a recommendation to use translated terms where possible / the requirement, as a precondition, that the word be one actually in wide use online and offline / Case 1: ``{\hangulfont 백과}'' (`encyclopedia') in {\hangulfont 위키백과} (`Wikipedia') has the same usage and meaning as the everyday word and is accepted naturally / Case 2: in ``{\hangulfont 바깥} {\hangulfont 고리}'' (`external link', literally `outside ring') both ``{\hangulfont 바깥}'' and ``{\hangulfont 고리}'' are far from their actual usage and meaning, so the translation is awkward / Case 2 given as a fitting example of why the precondition is needed. \\
Original & So I think this kind of one-sided, contrived translation practice should be avoided, and that we should use translated terms where possible but add the condition that the word is one in wide use on- and offline. In {\hangulfont 위키백과}, {\hangulfont 백과} has the same usage and meaning as the word {\hangulfont 백과} we all use, so it comes across naturally, but in {\hangulfont 바깥} {\hangulfont 고리} both words, {\hangulfont 바깥} and {\hangulfont 고리}, are far from the usage and meaning of the words we actually use, so it becomes an awkward translation; that would be a good example. \\
Own & I'll give {[}my view{]} to this discussion. / I think one-sided, contrived translation practice must be stamped out ({\hangulfont 척결되야}, a non-standard spelling, kept as generated). I recommend using translated terms, but I think the precondition must be that the word is one actually in wide use online and offline. For example, `{\hangulfont 백과}' in {\hangulfont 위키백과} is an everyday word with the same meaning, so it is accepted naturally, but `{\hangulfont 바깥}' and `{\hangulfont 고리}' in `{\hangulfont 바깥} {\hangulfont 고리}' both differ in meaning from everyday language, so it becomes an awkward translation; as this shows, I think there must be a precondition for using translated terms. \\
Next & One-sided and contrived translation practice {\hanjafont 悖} {[}a Chinese character, `runs against'{]} Wikipedia's purpose and instead leads readers to misunderstand. The recommendation to use translated terms where possible is because only widely used words and expressions can raise Wikipedia's accuracy and accessibility. The precondition for using translated terms is that the word is actually widely used or exists online and offline. / Case 1: ``{\hangulfont 백과}'' in {\hangulfont 위키백과} has the same usage and meaning as the everyday word and is accepted naturally. / Case 2: in ``{\hangulfont 바깥} {\hangulfont 고리}'', ``{\hangulfont 바깥}'' and ``{\hangulfont 고리}'' are both, from their actual usage and meaning, {[}a fragment{]} far away, ring (and {\hangulfont 고리} is somewhat like that too) a needless translation, so awkward. This case would be a fitting example of why the precondition is needed. \\\bottomrule
\end{tabularx}\caption{The worked example in English, under CC BY-SA 4.0. A slash between spaces marks a line break. Original is the gold version. Own and Next are the outputs of the own adapter and the next-author adapter.}\end{table}

\begin{table}[!htb]\centering\footnotesize\setlength{\tabcolsep}{5pt}\begin{tabular}{lrl}\toprule
Set & n & Used for \\\midrule
Main split, 876 same-page negatives & 1,010 triples & Verifier comparison \\
Second split & 736 triples & Verifier comparison \\
Bullet test, one work per author & 294 works & Table 1, Table 2 \\
Held-out reference, another text by the author & 267 works & Table 2 \\
Grid's chosen cell, five folds & 264 works & Training settings \\
LambdaG held-out texts, 8 held out, no degenerate output & 217 works & Table 2 \\
k curve, at least 32 other texts & 145 authors & Number of training texts \\
New authors, outside both splits & 174 authors & New-author check \\
English Wikipedia talk & 40 + 160 authors & English run \\
Literary corpus & 358 triples & Literary check \\\bottomrule
\end{tabular}\caption{Evaluation sets. Rows 4 to 7 are subsets of the 294 bullet-test works, fixed by the rule named after the set.}\end{table}

\FloatBarrier\section*{C\quad The M2 verifier}
\addcontentsline{toc}{section}{Appendix C: The M2 verifier}

\subsection*{C.1 Construction and scoring}
\addcontentsline{toc}{subsection}{C.1 Construction and scoring}

M2, the Korean verifier of Section 3.4, is Qwen3-8B with a LoRA adapter of rank 16. For each of five folds we train one adapter on 3,000 pairs of that fold's training authors. The prompt holds two texts, each cut to its first 460 Qwen3 tokens, and the target is the answer yes or no. The score of a pair is log p(yes)~− log p(no), averaged over both text orders. For every method we report the pooled ROC-AUC of all positive scores against all negative scores of the topic-matched triples. M2 scores are uncalibrated and lean to ``no'': they are above 0 on 39.1\% of the 2,020 pairs of the main split, so we claim no accuracy and no threshold.

The baselines use existing or public code: word-final frequency, character 1-4 gram TF-IDF, LambdaG on the original and on word-final-only text, bge-m3 \citep{chen-2024-bge-m3} and LUAR-MUD \citep{rivera-soto-2021-luar}, each run on the full text and on the first 460 tokens, which gives twelve variants. The planned comparison is M2 minus the best of these twelve on the evaluation triples, which is conservative for M2, and it passes when the lower bound of its interval is above 0. All intervals are 97.5\% paired bootstrap intervals over triples. The surface-feature verifier M1 is a logistic regression on eleven similarity features, trained on the same pairs as M2. Settings of the further trained and prompted systems are in ledger A.4.

\subsection*{C.2 Detection and its limits}
\addcontentsline{toc}{subsection}{C.2 Detection and its limits}

On the 1,010 triples of the main split, the best baseline is character TF-IDF on the first 460 tokens (Table A5). The public encoders and LambdaG stay near chance, far below the surface features. M2 is above the best baseline, above M1, and above a supervised character TF-IDF classifier trained on M2's training pairs. It is also above a contrastive encoder (ko-sroberta-multitask) fine-tuned on each fold's training authors, which has the highest AUC among the comparators of the table that use no LLM. M1 is not separated from the supervised character TF-IDF classifier (+0.0041 {[}−0.0185, +0.0286{]}), and we do not claim it as a contribution. The pooled AUC mixes the score scales of five separately trained adapters, but M2 is above the baseline in every fold, and an author-cluster bootstrap agrees. Per-fold values, PAN measures and the details of M1 are in ledger A.3, A.5 and A.6.

The prompted LLMs were added after the first results. Each sees the LIP prompt of \citet{huang-etal-2024-llm-authorship} and the same 2,020 pairs, and its score is its stated confidence, its yes/no answer, or for GPT-4.1 the log-probability of its answer token. M2 is above GPT-4.1 and is not separable from GPT-6.1 Sol. Claude Opus 5.5 scores above M2 in all three variants, although for the yes/no variant the author-cluster interval includes zero. Gemini 4 Argon is below chance on its raw confidence but above M2 when a no answer is scored as 1~− confidence. The comments are public Wikipedia text, so these models may have seen them in pre-training. Further variants are in ledger A.18.

\begin{table}[!htb]\centering\small\setlength{\tabcolsep}{5pt}\begin{tabular}{lrr}\toprule
Method & AUC & M2~− method \\\midrule
LambdaG & 0.5561 & n/a \\
LUAR-MUD & 0.5639 & n/a \\
bge-m3 & 0.5871 & n/a \\
Word-final frequency & 0.7170 & n/a \\
Char. 1-4 gram TF-IDF & 0.7273 & +0.0990 {[}+0.0783, +0.1223{]} † \\
M1, surface features & 0.7590 & +0.0673 {[}+0.0446, +0.0905{]} \\
M2, LoRA pair verifier & 0.8263 & n/a \\
Supervised char. TF-IDF ‡ & 0.7549 & +0.0714 {[}+0.0507, +0.0945{]} \\
Contrastive encoder, fine-tuned & 0.7913 & +0.0350 {[}+0.0159, +0.0528{]} \\
GPT-4.1, LIP prompt (English) ‡ & 0.6860 & +0.1402 {[}+0.1156, +0.1647{]} \\
GPT-6.1 Sol, confidence ‡ & 0.8203 & +0.0059 {[}−0.0165, +0.0286{]} \\
Gemini 4 Argon, confidence ‡ & 0.4609 & +0.3653 {[}+0.3364, +0.3940{]} \\
Gemini 4 Argon, answer-signed ‡ & 0.9176 & −0.0914 {[}−0.1090, −0.0740{]} \\
Claude Opus 5.5, yes/no ‡ & 0.8470 & −0.0208 {[}−0.0403, −0.0010{]} \\
Claude Opus 5.5, Korean confidence ‡ & 0.9165 & −0.0902 {[}−0.1068, −0.0737{]} \\
Claude Opus 5.5, confidence ‡ & 0.9201 & −0.0939 {[}−0.1103, −0.0776{]} \\\bottomrule
\end{tabular}\caption{Main split, 1,010 triples, folds pooled: ROC-AUC and M2 minus the method with its 97.5\% interval where computed. The contrastive encoder is outside the twelve baseline variants. † planned comparison (Appendix A). ‡ added after the first results.}\end{table}

The margin over the best baseline holds under a second training seed and on the second split, which has 371 authors from the forum and four talk namespaces (Table A6). On the literary corpus of the earlier diagnostic study the result reverses. M2, trained with the main setting, falls below word-final frequency on the full text, and the upper bound of the interval is below zero. Its training loss stays near the chance-level value at most logged steps, and two of five folds end there. Page classes of the second split, encoder classifiers and loss curves are in ledger A.2 and ledger A.3, and an earlier encoder study on the literary corpus is in ledger A.1.

\begin{table}[!htb]\centering\small\setlength{\tabcolsep}{5pt}\begin{tabular}{lrrrr}\toprule
Sample & n & M2 & Best baseline & M2~− best baseline \\\midrule
Main split, seed 1 † & 1,010 & 0.8341 & 0.7273 & +0.1068 {[}+0.0865, +0.1290{]} \\
Second split † & 736 & 0.7986 & 0.7298 & +0.0688 {[}+0.0385, +0.0988{]} \\
Literary corpus † & 358 & 0.6380 & 0.7026 & −0.0646 {[}−0.1095, −0.0184{]} \\\bottomrule
\end{tabular}\caption{Pooled ROC-AUC of M2 and of the best baseline on n triples, with the 97.5\% interval of the difference. Outside the main split the best baseline is word-final frequency on the full text. † planned comparison (Appendix A).}\end{table}

Three cues that a talk-page comment can carry do not explain the margin. M2 stays above the supervised character n-gram verifier on the triples without signature residue, on the triples where length does not favour the positive and without shared mentions, and above the best baseline on the residue-free triples. To locate what M2 reads, we score the second-split adapters without re-training on texts that show only word ends and on texts that hide them: both lose accuracy, M2 on word ends alone falls below the best surface baseline computed on the original text, and word ends alone keep more accuracy than the text without them in 5 of 5 folds, which suggests that the margin draws on word-final information, although the lower bound of that difference is +0.0001, at the boundary.

\FloatBarrier\section*{D\quad Reference texts: additional results}
\addcontentsline{toc}{section}{Appendix D: Reference texts: additional results}

\subsection*{D.1 The original as reference}
\addcontentsline{toc}{subsection}{D.1 The original as reference}

On the bullet-to-text test an adapter on the 8B model writes each author's text from a content bullet list. The test has 294 works, and none of the originals is among the adapters' training texts. Table A7 gives the single-reference results that qualify Section 4. With the original as the reference, M2 does not score the author's own text above the own adapter's output. The interval of this difference includes zero, and the difference is TOST-equivalent within ±0.10 of the fold's gap (the mean M2 difference between same-author and different-author pairs of the fold's test triples; ledger A.8), a margin that we fixed after the 95\% intervals and before the 90\% intervals. Scored against itself, the original gets no more than the output.

\begin{table}[!htb]\centering\footnotesize\setlength{\tabcolsep}{5pt}\begin{tabular}{llr}\toprule
Comparison & Reference & Difference {[}95\% interval{]} \\\midrule
Own~− original, M2 & original & +0.031 {[}−0.036, +0.094{]} \\
Own~− original scored against itself, M2 & original & +0.018 \\
Own~− untuned, chrF & original & −1.52 {[}−2.46, −0.67{]} \\
Own~− untuned, bullet-masked chrF & original & +2.43 {[}+1.75, +3.10{]} \\
Own~− original, M2 & 1 held-out text & −0.829 \\
Own~− original, 174 new authors, M2 & 1 held-out text & −0.178 \\
Own~− next-author, 174 new authors, M2 & 1 held-out text & +0.727 \\\bottomrule
\end{tabular}\caption{Differences of the own adapter with the original (294 works) or one held-out text (267 works; 174 new authors where stated) as the reference. On the 174 new authors the interval of the own adapter minus the original includes 0.}\end{table}

Against the original, M2 reads shared wording more than style. Copies of the original that keep fewer of its words score further below it, and M2 scores the original 0.285 below a copy with its sentences shuffled. With a held-out reference the interval of that last difference includes 0. These copies are in ledger A.8. Raw chrF mostly reads what the bullets carry, because the bullet lists already reach 27.91 chrF to the original, so we also drop the bullet words from both texts. Results by bullet overlap, for training-free systems and on copied spans are in ledger A.8 and A.11; the shared adapter is in ledger A.15.

\subsection*{D.2 LambdaG}
\addcontentsline{toc}{subsection}{D.2 LambdaG}

LambdaG \citep{nini-2026-lambdag} is a likelihood verifier that scores a text against known texts by the author. We run it on word-final-only text, in which each word is cut to its last two characters, also for the LambdaG rows of Table 2. With the adapter's training texts as known texts, LambdaG scores the own adapter above the original. With 8 held-out texts of the author as known texts, it scores the own adapter below the original on the same 217 works, and Table 1 gives both differences. We read the difference above the author as similarity to the training texts rather than exaggeration of the author, but we did not test this cause for LambdaG. The comparison of three training regimes on LambdaG is in ledger A.11.

\subsection*{D.3 Further inflation checks}
\addcontentsline{toc}{subsection}{D.3 Further inflation checks}

\textbf{Training settings.} Figure 3 compares adapters that differ in their training settings, and Table A8 gives its values. The default adapter has rank 8 and is trained with a learning rate of \mbox{1e-4} until the first epoch whose mean training loss is below 0.1 (the training-loss stop). The held-out stop keeps the epoch with the lowest loss on two held-out texts by the author, which at eight training texts is the first epoch for 239 of 264 authors. The three fixed budgets are 2 epochs, 3 epochs, and 3 epochs with rank 2 and dropout 0.1. Two further settings use a lower learning rate (\mbox{3e-5}) with early stopping on held-out loss, at rank 16 and at rank 32: the held-out loss is checked every 2 steps and the lowest check is kept. Memorisation is the share of outputs that repeat a 100-character span of a training text when the adapter is given its training bullets.

The held-out stop removes most of the inflation, but it is not a better adapter. With the original as the reference, its margin over the next-author adapter on M2 is +0.078 {[}−0.046, +0.205{]} (264 works), and with a held-out reference the margin is +0.145 {[}+0.038, +0.254{]} (262 works; ledger A.8). Its chrF to the original is −7.07 {[}−8.62, −5.52{]} relative to the untuned model, and 32.6\% of its outputs are degenerate, against 6.8\% at the training-loss stop. The fixed budgets, the learning-rate search and the results for 2 and 32 training texts are in ledger A.15.

\begin{table}[!htb]\centering\small\setlength{\tabcolsep}{5pt}\begin{tabular}{lrrr}\toprule
Training setting & n & II {[}95\% interval{]} & Memorisation \\\midrule
Training-loss stop & 266 & +0.38 {[}+0.25, +0.53{]} & 95.5\% \\
Held-out stop & 261 & +0.09 {[}−0.05, +0.24{]} & 0.9\% \\
Fixed 2 epochs & 261 & +0.38 {[}+0.23, +0.53{]} & 1.7\% \\
Fixed 3 epochs & 261 & +0.30 {[}+0.16, +0.44{]} & 24.6\% \\
Rank 2, dropout 0.1, 3 epochs & 261 & +0.41 {[}+0.27, +0.55{]} & 2.7\% \\
\mbox{3e-5} with early stopping, rank 16 & 261 & +0.21 {[}+0.06, +0.35{]} & 0.9\% \\
\mbox{3e-5} with early stopping, rank 32 & 261 & +0.14 {[}−0.01, +0.29{]} & 0.7\% \\
k = 32, training-loss stop & 144 & +0.13 {[}−0.05, +0.31{]} & 96.2\% \\
k = 32, held-out stop & 144 & +0.12 {[}−0.08, +0.31{]} & 2.4\% \\
Next-author adapter &  & +0.11 {[}−0.03, +0.25{]} & 0.1\% \\\bottomrule
\end{tabular}\caption{Inflation index (II) of the own adapter on Korean M2 by training setting, as plotted in Figure 3, with the share of outputs that reproduce a training text verbatim. n is the number of works. Settings use eight training texts unless noted. The two \mbox{3e-5} rows use a lower learning rate (\mbox{3e-5}) with early stopping on held-out loss. The last row is the next-author adapter at the training-loss stop (k = 8), for comparison.}\end{table}

Table A9 adds four checks to the inflation index of Table 1 and Figure 3. The first block repeats the training-loss stop at k = 8 with two further seeds and two further base models. The second divides II by the per-fold gaps of M2, because M2's scale varies by fold. The own adapter's II stays above zero in these units, and the other two systems are not separated from zero. Fold 4 has a negative self score, and leaving it out does not change the result.

\begin{table}[!htb]\centering\small\setlength{\tabcolsep}{5pt}\begin{tabular}{lrr}\toprule
Quantity & n & Value {[}95\% interval{]} \\\midrule
\multicolumn{3}{l}{\emph{II on M2, k = 8 training-loss stop}} \\
Seed 1 & 266 & +0.45 {[}+0.32, +0.60{]} \\
Seed 2 & 266 & +0.36 {[}+0.22, +0.51{]} \\
Gemma 3 12B & 266 & +0.31 {[}+0.18, +0.44{]} \\
Qwen3-32B & 266 & +0.37 {[}+0.23, +0.53{]} \\
\addlinespace[2pt]
\multicolumn{3}{l}{\emph{II on M2 in per-fold gaps}} \\
Own adapter & 266 & +0.20 {[}+0.13, +0.27{]} \\
Next-author adapter & 266 & +0.06 {[}−0.02, +0.13{]} \\
Untuned model & 266 & +0.02 {[}−0.06, +0.10{]} \\
Own adapter, without fold 4 & 216 & +0.20 {[}+0.12, +0.28{]} \\
\addlinespace[2pt]
\multicolumn{3}{l}{\emph{Prompted Claude Opus 5.5 as verifier}} \\
II, own adapter & 267 & +0.104 {[}+0.076, +0.132{]} \\
II, next-author adapter & 267 & +0.018 {[}−0.010, +0.047{]} \\
II, untuned model & 267 & +0.025 {[}−0.002, +0.051{]} \\
Original~− own, training reference & 267 & +0.122 {[}+0.097, +0.147{]} \\
\addlinespace[2pt]
\multicolumn{3}{l}{\emph{Topic of the references, bge-m3}} \\
Cosine difference & 266 & −0.004 {[}−0.014, +0.006{]} \\
ρ, cosine difference and II & 266 & −0.06 {[}−0.18, +0.06{]} \\
ρ, II and similarity, topic partialled out & 266 & +0.196 {[}+0.076, +0.314{]} \\\bottomrule
\end{tabular}\caption{Further checks of the own adapter's inflation index (II): other seeds and base models, II in per-fold gaps, Claude Opus as the verifier, and topic similarity of the references to the original. n is the number of works.}\end{table}

\textbf{Claude Opus as verifier.} Prompted Claude Opus 5.5 has the highest ROC-AUC among our verifiers, so we repeat the reference comparison with it. Opus also inflates the own adapter when it is given a training text, while the next-author adapter and the untuned model are not separated from zero. Against the training reference the original stays above the own adapter. We newly scored only the training reference, and the held-out side reuses earlier Opus scores. Opus may have been pre-trained on these comments (Appendix G).

\textbf{Topic.} A training reference could raise the score because it is closer in topic to the original. In bge-m3 embeddings the training reference is no closer to the original than the held-out references, and the topic difference does not track II. The correlation of II with the similarity measure of Section 4 is nearly unchanged when the topic difference is partialled out. Overlap of the originals with training texts and repetition of training texts by setting are in ledger A.11.

\subsection*{D.4 Selection by held-out score}
\addcontentsline{toc}{subsection}{D.4 Selection by held-out score}

The held-out score of Section 4 can serve as a selection rule that never reads the original. Per author we choose among seven of its k = 8 adapters: the two stops, three fixed budgets and two lower-learning-rate settings of Table A8. Two filters drop outputs that miss their content bullets and outputs that copy a training text, and we take the highest remaining M2 against the held-out mean. Table A10 shows that the rule beats the training-loss stop alone. Without the two filters, the highest-M2 pick moves away from the original, with raw chrF −2.68 {[}−3.86, −1.49{]}.

\begin{table}[!htb]\centering\footnotesize\setlength{\tabcolsep}{5pt}\begin{tabular}{llrr}\toprule
Comparison & Score & n & Difference {[}95\% interval{]} \\\midrule
Seven adapters~− training-loss stop & Bullet-masked chrF & 261 & +0.88 {[}+0.29, +1.47{]} \\
Seven adapters~− training-loss stop & Raw chrF & 261 & +1.58 \\
Eight samples~− greedy output & Bullet-masked chrF & 266 & +1.12 {[}+0.59, +1.66{]} \\
Preference-tuned pilot & Bullet-masked chrF & 50 & −0.52 {[}−1.97, +0.80{]} \\\bottomrule
\end{tabular}\caption{Closeness to the original of outputs chosen by the selection rule, relative to the training-loss stop's greedy output, and the gain of the preference-tuning pilot. n is the number of works.}\end{table}

The same choice among eight outputs sampled from the training-loss stop at temperature 0.7 also beats its greedy output. In 19 works no sample passes the filters and the greedy output stays. We also preference-tuned each adapter on the rule's own choices, using DPO on the chosen sample against the other samples. The pairs are built without the original. We tried this on a 50-work pilot and did not extend it, because its mean gain on bullet-masked chrF was not above zero.

\subsection*{D.5 AuthorMix rerun}
\addcontentsline{toc}{subsection}{D.5 AuthorMix rerun}

We reran a scaled-down AuthorMix with the settings of Table A11. It mixes ten high-resource LoRA adapters layer by layer, and its gradient-free search scores outputs against the 16 sentences X\_t of each target author. The inflation index is STAR Toward against X\_t minus Toward against 16 sentences from another book of the same author. Our adapters train far less than the original ones. Validation loss was lowest at step 50 for all ten adapters and rose by step 300, where it was still below the base model's. The rule keeps the last adapter, so the mix starts from overfitted adapters.

\begin{table}[!htb]\centering\footnotesize\setlength{\tabcolsep}{5pt}\begin{tabular}{ll}\toprule
Setting or result & Value \\\midrule
Base model & Llama-3.1-8B-Instruct, 4-bit \\
Adapter training, original & 5,000 steps, effective batch 32 \\
Adapter training, rerun & 300 steps at batch 8 on about 600 pairs (about four epochs) \\
Validation loss at step 50 & 1.439--1.949 \\
Validation loss at step 300 & 2.221--2.928 \\
Validation loss, base model & 3.995--4.602 \\
II difference, STAR † & −0.0008 {[}−0.0025, +0.0009{]} \\
II difference, LUAR & +0.0005 {[}−0.0011, +0.0022{]} \\
II difference, Style-Embedding & −0.0013 {[}−0.0042, +0.0014{]} \\\bottomrule
\end{tabular}\caption{Settings of the AuthorMix rerun and the inflation index of AuthorMix minus that of its base model over 10 target authors, with 95\% intervals. † planned comparison outside the 32 of Appendix A (pre-registered endpoint).}\end{table}

The primary endpoint is not met. Scored against the 16 sentences that AuthorMix optimises on, its Toward does not exceed the held-out score by more than the base model's does at alpha = 0.05. The exact sign-flip test gives p = 0.8086 over the 10 targets. The contrast on LUAR and Style-Embedding is an estimate only, without a test, and the full list of systems is in ledger A.11.

\FloatBarrier\section*{E\quad Bullet-to-text: additional results}
\addcontentsline{toc}{section}{Appendix E: Bullet-to-text: additional results}

\subsection*{E.1 Main contrasts and references}
\addcontentsline{toc}{subsection}{E.1 Main contrasts and references}

This appendix supports Section 5. Values are means over works with work-level 95\% bootstrap intervals from 2,000 draws. In Table A12, both daggered comparisons are positive with the original as the reference text, and the M2 margin over the next-author adapter holds at two further adapter seeds. Per seed, chrF separates the own adapter from the next-author adapter at one of three seeds and ending cosine at none (ledger A.8). The own adapter is not separated from the original on M2, and its chrF falls below the untuned model's. In a sentence-level audit, 25 of the own adapter's outputs and 25 of the next-author adapter's are truncated or repeat a sentence. The untuned model has 2.

The held-out reference is one other text by the author that appears in no adapter training and in no M2 training or validation pair, and 267 of the 294 authors have one. With it the own adapter still scores above the other three systems but below the original. On 174 authors outside both splits it is above the next-author adapter and not separated from the original. Prompted Claude Opus 5.5, which may have seen these comments in pre-training, scores the original above the own adapter by +0.679 with the original as the reference and by +0.226 with the held-out reference. Ledger A.8 gives fold values, calibrated scales and the held-out-stop adapters.

\begin{table}[!htb]\centering\small\setlength{\tabcolsep}{5pt}\begin{tabular}{llrr}\toprule
Own adapter minus & Score & n & Difference {[}95\% interval{]} \\\midrule
\multicolumn{4}{l}{\emph{Original as reference}} \\
Next-author adapter † & M2 & 294 & +0.290 {[}+0.208, +0.377{]} \\
Prompted model † & M2 & 294 & +0.798 {[}+0.727, +0.867{]} \\
Original & M2 & 294 & +0.031 {[}−0.036, +0.094{]} \\
Untuned model & chrF & 294 & −1.52 {[}−2.46, −0.67{]} \\
\addlinespace[2pt]
\multicolumn{4}{l}{\emph{Next-author adapter, by seed}} \\
Seed 1 & M2 & 294 & +0.205 {[}+0.130, +0.286{]} \\
Seed 2 & M2 & 294 & +0.232 {[}+0.168, +0.300{]} \\
Seed average per work & M2 & 294 & +0.242 {[}+0.190, +0.295{]} \\
Seed average per work & chrF & 294 & +0.99 {[}+0.17, +1.83{]} \\
Seed average per work & ending cosine & 294 & +0.013 {[}+0.000, +0.027{]} \\
\addlinespace[2pt]
\multicolumn{4}{l}{\emph{Held-out reference}} \\
Next-author adapter & M2 & 267 & +0.600 {[}+0.456, +0.746{]} \\
Original & M2 & 267 & −0.829 {[}−0.962, −0.694{]} \\
Untuned model & M2 & 267 & +1.157 {[}+1.022, +1.297{]} \\
Prompted model & M2 & 267 & +1.278 {[}+1.141, +1.417{]} \\
\addlinespace[2pt]
\multicolumn{4}{l}{\emph{Held-out reference, new authors}} \\
Next-author adapter & M2 & 174 & +0.727 {[}+0.541, +0.925{]} \\
Original & M2 & 174 & −0.178 {[}−0.371, +0.017{]} \\
\addlinespace[2pt]
\multicolumn{4}{l}{\emph{Next-author adapter, Opus as verifier}} \\
Held-out reference & Opus & 267 & +0.088 {[}+0.063, +0.112{]} \\
Original as reference & Opus & 294 & +0.015 {[}+0.001, +0.031{]} \\\bottomrule
\end{tabular}\caption{Differences of the own adapter from other systems, 95\% intervals. Seed 0 is the first row. † planned comparison (Appendix A).}\end{table}

\subsection*{E.2 Explanations of the margin}
\addcontentsline{toc}{subsection}{E.2 Explanations of the margin}

Table A13 collects checks on the margin of the own adapter over the next-author adapter. Masking the character 8-grams that an output shares with its adapter's training texts leaves the margin positive, and the control masks the same share of the next-author output at random positions. Masking the 8-grams that each scored text shares with its reference lowers the inflation index of the own adapter, which is +0.52 unmasked on the same 267 works, and leaves it above zero. Where the next author's training texts are at least as close to the target in topic, the margin holds by bge-m3 cosine, but the two adapters are not separated by TF-IDF cosine over word stems, so topic can explain part of the margin.

A supervised character n-gram verifier, trained per fold on the same pairs as M2, has an AUC of 0.6915 on the split's triples and also scores the own adapter above the next-author adapter. Against the original, M2 scores a copy lower the fewer of the original's words it keeps, so it reads shared wording as well as style. GPT-4.1 and GPT-6.1 Sol, each prompted with 8 of the author's texts, are closer to the original than the own adapter on chrF, bullet-masked chrF and ending cosine, and on M2 the own adapter is above GPT-4.1 but not separated from Sol. Sol repeats the original beyond the bullets in 13 works and may have seen it in pre-training, but it stays closer in chrF without the 16 span-sharing works. Further masks, topic-matched authors, and spliced, shuffled and paraphrased copies are in ledger A.8, and a replication of STYLL is in ledger A.8.

\begin{table}[!htb]\centering\footnotesize\setlength{\tabcolsep}{5pt}\begin{tabular}{llrr}\toprule
Check & Contrast & n & Difference {[}95\% interval{]} \\\midrule
\multicolumn{4}{l}{\emph{Masking copied spans, M2}} \\
8-grams shared with training texts & own~− next & 294 & +0.205 {[}+0.125, +0.284{]} \\
Next output at random, same share & own~− next & 294 & +0.275 {[}+0.193, +0.361{]} \\
8-grams shared with reference & II of own & 267 & +0.43 {[}+0.28, +0.58{]} \\
Change of II under this mask & II of own & 267 & −0.09 {[}−0.15, −0.02{]} \\
\addlinespace[2pt]
\multicolumn{4}{l}{\emph{Topic, M2}} \\
Next as close, bge-m3 & own~− next & 71 & +0.233 {[}+0.077, +0.378{]} \\
Next as close, word stems & own~− next & 38 & +0.046 {[}−0.169, +0.227{]} \\
\addlinespace[2pt]
\multicolumn{4}{l}{\emph{Character n-gram verifier}} \\
Outputs as written & own~− next &  & +0.252 {[}+0.143, +0.360{]} \\
Outputs as written & original~− own &  & +5.195 {[}+5.089, +5.307{]} \\
\addlinespace[2pt]
\multicolumn{4}{l}{\emph{Partial copies, M2}} \\
First 75\% kept, original as reference & copy~− original & 294 & −0.065 {[}−0.089, −0.043{]} \\
First 50\% kept, original as reference & copy~− original & 294 & −0.100 {[}−0.137, −0.066{]} \\
First 25\% kept, original as reference & copy~− original & 294 & −0.180 {[}−0.229, −0.133{]} \\
\addlinespace[2pt]
\multicolumn{4}{l}{\emph{Other systems, original as reference}} \\
Prompted GPT-4.1, M2 & own~− GPT-4.1 & 294 & +0.367 {[}+0.303, +0.432{]} \\
Prompted GPT-4.1, chrF & own~− GPT-4.1 & 294 & −4.27 {[}−5.11, −3.50{]} \\
Prompted GPT-4.1, bullet-masked chrF & own~− GPT-4.1 & 294 & −3.37 {[}−4.03, −2.76{]} \\
Prompted GPT-4.1, ending cosine & own~− GPT-4.1 & 294 & −0.059 {[}−0.075, −0.045{]} \\
Prompted GPT-6.1 Sol, M2 & own~− Sol & 294 & −0.012 {[}−0.068, +0.043{]} \\
Prompted GPT-6.1 Sol, chrF & own~− Sol & 294 & −9.45 {[}−10.41, −8.65{]} \\
Prompted GPT-6.1 Sol, bullet-masked chrF & own~− Sol & 294 & −8.61 {[}−9.38, −7.89{]} \\
Prompted GPT-6.1 Sol, ending cosine & own~− Sol & 294 & −0.100 {[}−0.117, −0.086{]} \\\bottomrule
\end{tabular}\caption{Checks on the margin of the own adapter. Copies that keep the first 75\%, 50\% or 25\% of the original's words, with the rest taken from the next-author adapter's output, score lower the less they keep. Next as close: the next author's training texts are at least as close in topic to the target. The n-gram verifier has its own scale. In the last block, a negative difference on chrF or ending cosine means that the prompted system is closer to the original than the own adapter.}\end{table}

\subsection*{E.3 LLM judges}
\addcontentsline{toc}{subsection}{E.3 LLM judges}

No human evaluation has been run. LLM judges answered the items of the designed rating task (Appendix H), and they do not stand in for it. Shown three of the author's training texts, Claude Opus 5.5, Sonnet 5.5 and Haiku 5.5 answer the 165 items of the task. They pick the own adapter over the next-author adapter in 74.3\% of the judgments of that comparison and the original over the own adapter in 80.0\% of the judgments of that one. Table A14 repeats the task on 147 items with three texts by the author that were used in no training. The preference for the own adapter over the next-author adapter falls but still holds, which we read as recognition of training texts plus style. The Claude judges come from the company whose model wrote the bullets, and all judges may have seen these comments in pre-training. Ledger A.8 gives per-judge results.

\begin{table}[!htb]\centering\scriptsize\setlength{\tabcolsep}{5pt}\begin{tabular}{lrrr}\toprule
Judges and comparison & Held-out shown & Training shown & Held-out~− training {[}95\% interval{]} \\\midrule
Claude, own over next & 63.4\% & 75.4\% & −12.0 {[}−20.3, −4.3{]} \\
Claude, own over untuned & 86.2\% & 86.6\% &  \\
Claude, original over own & 89.1\% & 78.7\% & own pick −10.5 {[}−15.9, −5.0{]} \\
Claude, original, unmarked & 70.7\% {[}56.4\%, 84.2\%{]} & 47.4\% &  \\
GPT-6.1 Sol, own over next & 70.7\% & 85.9\% & −15.2 {[}−26.1, −5.4{]} \\
Gemini 4 Argon, own over next & 73.9\% & 77.2\% & −3.3 {[}−14.1, +6.5{]} \\\bottomrule
\end{tabular}\caption{Shares of judgments, per comparison, among 147 items, that pick the first-named candidate when judges see held-out or training texts by the author. In the third row the difference is that of the share that picks the own adapter. Unmarked: judgments that do not mark the output as not making sense.}\end{table}

\FloatBarrier\section*{F\quad Other base models and English}
\addcontentsline{toc}{section}{Appendix F: Other base models and English}

\subsection*{F.1 Other base models}
\addcontentsline{toc}{subsection}{F.1 Other base models}

We re-run the bullet-to-text conditions of Section 5 on Gemma 3 12B and Qwen3-32B, both quantised to 4-bit, on all five folds, which gives 294 works. The last 16 layers are adapted on every model, and the LoRA cell, prompts, decoding and scoring are unchanged. M2 is the same Qwen3-8B verifier for every base model. Table A15 gives the M2 differences. The own adapter is above the next-author adapter, the prompted model and the untuned model on all three base models. The original is not separated from any own adapter on M2.

For the difference from the next-author adapter, the lower bound stays above zero on all five folds for Gemma 3 12B, on four folds for Qwen3-32B and on three folds for Qwen3-8B. With held-out known texts, LambdaG places the Qwen3-8B adapter below the author, and we did not re-score the other two models. Ledger A.12 has the settings, the single-fold values, a check of output form and the LambdaG differences.

\begin{table}[!htb]\centering\footnotesize\setlength{\tabcolsep}{5pt}\begin{tabular}{lrrr}\toprule
Own adapter minus & Gemma 3 12B & Qwen3-32B & Qwen3-8B \\\midrule
Next-author adapter † & +0.339 {[}+0.242, +0.438{]} & +0.233 {[}+0.155, +0.315{]} & +0.290 {[}+0.208, +0.377{]} \\
Prompted model † & +0.434 {[}+0.360, +0.508{]} & +0.610 {[}+0.538, +0.683{]} & +0.798 {[}+0.727, +0.867{]} \\
Untuned model & +0.551 {[}+0.477, +0.627{]} & +0.653 {[}+0.579, +0.727{]} & +0.376 {[}+0.308, +0.441{]} \\
Original & −0.047 {[}−0.127, +0.028{]} & −0.040 {[}−0.108, +0.028{]} & +0.031 {[}−0.036, +0.094{]} \\\bottomrule
\end{tabular}\caption{M2 difference of the own adapter from each other system and from the original, by base model, on 294 works, with 95\% intervals. † planned comparison.}\end{table}

\subsection*{F.2 English data and EN-M2}
\addcontentsline{toc}{subsection}{F.2 English data and EN-M2}

The data are the ConvoKit Wikipedia Talk Pages Corpus, licensed CC BY-SA 4.0. The corpus has no age metadata, so minors cannot be screened out. It has no page titles, so a work is one author's comments under one conversation root. We exclude bots, IP addresses and unknown users, remove markup and signatures, and drop works shorter than 250 characters, which leaves 110,058 works. The first check draws 40 authors with seed 0 and 6 targets of 250 to 1,000 characters per author. Each adapter is trained on 8 exemplars. Every text is used as its first line-bounded segment of at most 600 characters, and a single longer line is kept whole, so 38 of the 240 targets are shortened.

We planned the expansion after seeing the 40-author results and before any selection, training or scoring. It draws 160 further authors with seed 1 from the 837 eligible authors outside the 40, and takes 6 targets from each. All 160 adapters meet the training-loss stop. EN-M2 applies the Korean M2 recipe to English pairs, with 3,000 training pairs and 1,000 test pairs by authors outside the 200. Its test ROC-AUC is 0.90 {[}0.8808, 0.9173{]}, above the required lower bound of 0.70. Its test is easier than that of the Korean M2. Of the 500 different-author test pairs, 480 (96\%) could not be matched on the conversation and so differ in topic too, and against the 20 matched pairs alone the ROC-AUC is 0.7671.

\subsection*{F.3 English results}
\addcontentsline{toc}{subsection}{F.3 English results}

A first check on 40 authors and 240 targets used LUAR only, because M2 had no English version at the time. Its two planned comparisons, the own adapter above the next-author adapter and above the prompted model on LUAR, are positive and are listed in Appendix A. We planned the 160-author expansion after seeing these results, and the details of the first check are in ledger A.13.

The 160-author expansion is in Table A16. The two comparisons marked §, on LUAR and on EN-M2 with the original as the reference, are both positive. With a held-out text of the author as EN-M2's reference, the own adapter stays above the next-author adapter but falls below the original, as in Korean. With authors resampled, two differences are not separated from zero: LambdaG against the original with held-out known texts, and chrF against the next-author adapter. Ledger A.13 has the differences from the untuned model, the pooled estimate, and the copying and truncation counts.

\begin{table}[!htb]\centering\small\setlength{\tabcolsep}{5pt}\begin{tabular}{llrr}\toprule
Own adapter minus & Scorer & Difference {[}95\% CI{]} & Authors resampled \\\midrule
Next-author adapter § & LUAR & +0.0327 {[}+0.0259, +0.0392{]} & {[}+0.0221, +0.0443{]} \\
Next-author adapter § & EN-M2 & +0.4555 {[}+0.4018, +0.5100{]} & {[}+0.3641, +0.5518{]} \\
Next-author adapter & EN-M2 (h) & +0.6104 {[}+0.5559, +0.6689{]} &  \\
Next-author adapter & chrF & +1.30 {[}+0.36, +2.20{]} & {[}−0.16, +2.96{]} \\
Prompted model & LUAR & +0.0139 {[}+0.0072, +0.0205{]} &  \\
Prompted model & EN-M2 & +0.5801 {[}+0.5327, +0.6235{]} &  \\
Original & LUAR & +0.0002 {[}−0.0059, +0.0062{]} &  \\
Original & EN-M2 (h) & −0.2335 {[}−0.2862, −0.1810{]} &  \\
Original & LambdaG (h) & +13.50 {[}+0.63, +29.06{]} & {[}−0.53, +38.08{]} \\\bottomrule
\end{tabular}\caption{English expansion, 160 authors and 960 targets, with target-level 95\% intervals. (h): held-out reference text or held-out known texts. § planned comparison outside the 32 of Appendix A.}\end{table}

\subsection*{F.4 TinyStyler and Style-Embedding}
\addcontentsline{toc}{subsection}{F.4 TinyStyler and Style-Embedding}

We compare the own adapter with TinyStyler \citep{horvitz-2024-tinystyler} on the 160 authors. Table A17 gives the differences and the inflation index of Section 3.3 on Style-Embedding. The own adapter is above TinyStyler on LUAR and EN-M2 and below it on Style-Embedding, the embedding that TinyStyler conditions on. On LUAR, TinyStyler is not separated from the untuned model, and with a held-out reference text EN-M2 places it below the original. The inflation index on Style-Embedding is above zero for TinyStyler. The interval of the own adapter includes 0, and so do the intervals of the other four systems. The ancillary file S1 has the full readings.

\begin{table}[!htb]\centering\small\setlength{\tabcolsep}{5pt}\begin{tabular}{llr}\toprule
Quantity & Scorer & Value {[}95\% CI{]} \\\midrule
\multicolumn{3}{l}{\emph{Difference}} \\
Own adapter~− TinyStyler & LUAR & +0.0164 {[}+0.0101, +0.0229{]} \\
Own adapter~− TinyStyler & EN-M2 & +0.740 {[}+0.686, +0.790{]} \\
Own adapter~− TinyStyler & Style-Embedding & −0.0780 {[}−0.0911, −0.0647{]} \\
TinyStyler~− untuned model & LUAR & −0.0031 {[}−0.0084, +0.0022{]} \\
TinyStyler~− original & EN-M2 (h) & −0.794 {[}−0.855, −0.736{]} \\
\addlinespace[2pt]
\multicolumn{3}{l}{\emph{Inflation index (II)}} \\
TinyStyler & Style-Embedding & +0.0254 {[}+0.0104, +0.0412{]} \\
Own adapter & Style-Embedding & +0.0077 {[}−0.0067, +0.0227{]} \\\bottomrule
\end{tabular}\caption{TinyStyler on the 160 English authors: differences between systems, and the inflation index on Style-Embedding. (h): held-out reference text.}\end{table}

\FloatBarrier\section*{G\quad Additional limitations}
\addcontentsline{toc}{section}{Appendix G: Additional limitations}

\textbf{Bullets and tool definitions.} Part of the Claude Sonnet 5.5 calls carried the tool definitions of account connectors in the prompt, among them 14.6\% of the Korean bullet lists. We kept the bullets as written. Table A18 shows that the main result holds without the 41 affected works: the own adapter stays above the next-author adapter on M2, and its II is +0.42. The margin over the next-author adapter also holds with test bullets written by GPT-4.1, which hold 13.8\% of the original's words against 30.6\% for the Claude Sonnet 5.5 bullets.

\begin{table}[!htb]\centering\small\setlength{\tabcolsep}{5pt}\begin{tabular}{lrr}\toprule
Check & n & Value {[}95\% interval{]} \\\midrule
\multicolumn{3}{l}{\emph{Own~− next-author adapter on M2}} \\
Without works with tool definitions & 253 & +0.305 {[}+0.219, +0.397{]} \\
GPT-4.1 test bullets, seed-1 adapters & 294 & +0.304 {[}+0.210, +0.394{]} \\
Claude Sonnet 5.5 test bullets, seed-1 adapters & 294 & +0.205 {[}+0.130, +0.286{]} \\
\addlinespace[2pt]
\multicolumn{3}{l}{\emph{Membership AUC, untuned Qwen3-8B}} \\
Min-K\%++, comments after every release & 294 & 0.548 {[}0.501, 0.594{]} \\
Min-K\%++, January 2024 to first release & 294 & 0.533 {[}0.486, 0.580{]} \\
Min-K\% & 294 & 0.542 {[}0.497, 0.591{]} \\
Mean log-probability & 294 & 0.445 {[}0.400, 0.492{]} \\
\addlinespace[2pt]
\multicolumn{3}{l}{\emph{Min-K\%++ difference within authors}} \\
Earlier~− later texts & 99 & +0.14 {[}−0.23, +0.52{]} \\\bottomrule
\end{tabular}\caption{Checks behind the limitations. n counts works, originals (separated from 300 comparison comments) or authors. The first row leaves out the 41 works whose bullets carried tool definitions.}\end{table}

\textbf{Verifier, data and baselines.} The score of M2 for a text against itself varies across folds, with fold means from −0.314 to 2.716, which is why we take differences within a work under that fold's verifier. M2 does not transfer to every domain. On a literary corpus of 358 triples it scores below word-final frequency, 0.6380 against 0.7026 pooled over all five folds, and it is also below on 4 of 5 folds taken separately. Both splits come from one site and share its register. In the second split, 139 of 736 triples carry a substituted-template warning text or wiki markup residue, because the namespace build did not strip them. Removing these triples does not change the direction of the margin (ancillary file S1).

The decision rule takes as best baseline the highest of the twelve baselines of Appendix C on the test triples, which favours the baseline. We added the supervised surface-feature verifier and the GPT-4.1 prompting baseline after seeing the first results on fold 0, and later three further LLM verifiers, of which Claude Opus 5.5 scores above M2 (Appendix C). All three may have seen these public comments in pre-training. Results for variants that are not in this edition, such as a verifier retrained on machine imitations and other numbers of training texts, are in ledger A.14 and A.15, and their limitations in ledger A.9.

\textbf{Memorisation, exposure and alternative explanations.} Given each of its up to 8 training bullet lists at temperature 0, the own adapter returns at least 100 consecutive characters of the training text in 95.5\% of the cases and the whole text word for word in 46.4\%. The next-author adapter almost never does. Where the next author's training texts are at least as close to the target in word-stem topic similarity, own minus next is +0.046 with an interval that includes 0, so topical wording may explain part of the difference. The masked re-scoring removes only shared character 8-grams.

Of the 294 works, 287 predate the first model release, so we applied membership scores to the untuned Qwen3-8B \citep{zhang-2025-minkpp}. Min-K\%++ rates the originals above comments written after every release, and the lower bound of its AUC is 0.501, which indicates exposure or a date shift. Against comments from January 2024 to the first release, Min-K\%++ is not separated from chance. Min-K\% is not separated from chance either, and mean log-probability is below chance, so the three scores disagree in direction.

\textbf{Reference texts, models and sensitivity.} The held-out mean has four texts for 242 of the 266 Korean works and one to three for the other 24. Seeds 1--2, Gemma 3 12B and Qwen3-32B show the same Korean inflation, with II from +0.31 to +0.45, but the two other base models repeat only part of Table 2. At k = 8, 5.8\% of the outputs of the Qwen3-8B own adapter contain Japanese kana, against 0.3\% for the untuned model, and with the held-out stop the share is 0.0\%. Against the untuned model, chrF and ending cosine could miss gains below 1.29 and 0.023 at power 0.8.

\textbf{Statistics.} We read each of the 32 comparisons of Appendix A on its own paired bootstrap interval. No rule corrects across the 32, so the chance of at least one false positive among them is not held at 5\%. A comparison that is not separated is read as inconclusive, and no rule sets a minimum effect size. The Holm and Benjamini--Hochberg check added afterwards keeps all 28 separated comparisons separated. The two English-expansion comparisons form a separate family outside the 32 and outside this check. The rules were committed to a private repository, whose heads we timestamped with OpenTimestamps on \mbox{2026-10-25} and on \mbox{2026-11-13}.

\FloatBarrier\section*{H\quad Extended ethics statement}
\addcontentsline{toc}{section}{Appendix H: Extended ethics statement}

\textbf{Licences.} Every benchmark text is a public Korean Wikipedia discussion comment under CC BY-SA 4.0, from the community forum archive and, in the second split, four talk namespaces of the same site. The English talk check uses the ConvoKit Wikipedia Talk Pages Corpus under the same licence. The literary corpus of the diagnostic study holds Korean Wikisource works that are public domain or under KOGL type 1, CC BY or CC0, and the scenes of a Hugging Face set of 1930s novels under the MIT licence. The English control is the Blog Authorship Corpus \citep{schler-2006-blog}, which its authors distribute for non-commercial research use without a redistribution clause.

\textbf{What is released.} We do not redistribute texts. For the Wikipedia data we will release page titles, revision ids, work and author ids, SHA-1 hashes of each work text and a script that fetches the exact revisions from the public API and rebuilds the corpus byte for byte. For the bullet-to-text test we will release the pair index files, the prompts, the generation settings and the API ledgers. The LLM outputs, which are the normalisations, the bullets and the adapter and baseline outputs, are withheld until we settle whether the terms of the models that produced them allow redistribution. The literary rebuild is not pinned to revision ids. Per-corpus details are in ledger A.10.

\textbf{Authors under 18.} The Blog Authorship Corpus consists of blogger.com posts collected in 2004 and includes authors aged 13 to 17. We use 2,492 posts of 500 authors sampled without regard to age, and 227 of them are 13 to 17. The main reading of the control holds without them. On the 323 triples whose three texts are all by adults, LUAR-MUD minus character n-grams is +0.1896 {[}+0.1531, +0.2232{]} at the 97.5\% level. The value for all 998 triples is +0.1582, and a new sample drawn only from adults gives the same reading (ledger A.10). The control supports no claim of the paper, and a reuse of it should exclude authors under 18 by default. The released copies replace the ids of these authors with keyed hashes. Korean Wikipedia and the ConvoKit corpus record no age, so comments by minors cannot be screened out there.

\textbf{Third-party processing.} Forum comment texts were sent to commercial LLM services, and Table A19 lists each model, the route and the task. The judges received author texts, bullets, outputs and originals with user names masked. The namespace works and the literary corpus were processed locally only, as were the per-author adapters, M2 and every discriminator. The Claude calls ran through a subscription and fall under Anthropic's consumer terms, not its commercial API terms. We have not reviewed whether bulk research generation through a subscription agrees with those terms, which is one reason the bullets are not released.

\begin{table}[!htb]\centering\small\setlength{\tabcolsep}{5pt}\begin{tabularx}{\textwidth}{ll>{\raggedright\arraybackslash}X}\toprule
Model & Route & Task \\\midrule
GPT-4.1 & OpenRouter API & Zero-shot verification baseline, spelling normalisation, a second set of test bullets, STYLL and in-context baselines \\
GPT-6.1 Sol & OpenRouter API & Judge and one more in-context baseline \\
Gemini 4 Argon & OpenRouter API & Judge \\
Claude Sonnet 5.5 & Claude Code CLI & Bullets, written from the original comments and author texts of the evaluation authors, and judge on the 165 human rating items \\
Claude Opus 5.5 & Claude Code CLI & Judge on the 165 human rating items and verifier on both pairs per triple, on Table 2 outputs and in training-reference rescoring \\
Claude Haiku 5.5 & Claude Code CLI & Judge on the 165 human rating items \\\bottomrule
\end{tabularx}\caption{Third-party models that received comment texts, the route by which the texts were sent, and the task.}\end{table}

\textbf{Identification risk.} The task is authorship verification, so a trained verifier can link texts written under different accounts or outside Wikipedia to a Wikipedia user. User names are public on Wikipedia and recoverable from the rebuilt data. The author ids in our repository are unsalted SHA-1 prefixes of user names, which a list of user names can reverse. The released copies use keyed hashes, computed with HMAC-SHA256 under a key that is not released. This stops the dictionary reversal but not the linkage through the rebuilt data, so we count ids as a risk, not a protection. We release no trained verifier weights and no author-level predictions. Contributors can remove comments through normal Wikipedia processes or ask us to remove their author id from the released index files.

\textbf{Adapters and dual use.} An adapter trained on 2 to 32 texts of one pseudonymous contributor is a model of that person's surface habits and can impersonate that contributor. Given its own training bullets, an adapter returns the whole training text word for word in 46.4\% of the cases, so its weights carry that person's texts. We draw three lines. First, evaluation is against the author's original: no public AI-text detector or authorship verifier is used in training or as a reward, and M2 is reported, not optimised against. Second, we release no adapter weights and no author-level outputs. The released code and index files still let others train the same adapters from the public revisions, so this residual risk to pseudonymous editors remains. Third, the intended use is checking how style adaptation is scored. Neither the verifier nor the adapters are intended for de-anonymising individuals, for forensic attribution without legal process, for impersonating a contributor, for presenting machine-written text as human where disclosure is required, or for evading authorship verification or AI-text detection.

\textbf{Human evaluation and compute.} No human evaluation had been run at the time of writing. We designed a rating task for the bullet-to-text outputs, with its rater inclusion rules and decision rules fixed in a plan committed before any response, but did not run it. Raters would see public forum comments and generated text with user names masked. The task is not a premise of any claim, and the LLM judges that answered the same items do not stand in for it. We sought no institutional ethics review, because the work uses public texts only and involved no human participants. Local training and generation ran on one desktop machine, an Apple M5 Max running MLX, with 4-bit 8B to 32B models and LoRA adapters.

\FloatBarrier\section*{I\quad Responsible NLP checklist}
\addcontentsline{toc}{section}{Appendix I: Responsible NLP checklist}

\begin{table}[!htb]\centering\small\setlength{\tabcolsep}{5pt}\begin{tabularx}{\textwidth}{ll>{\raggedright\arraybackslash}X}\toprule
Item & Answer & Where \\\midrule
A1 Limitations & Yes & Limitations section, Appendix G \\
A2 Risks & Yes & Ethics statement, Appendix H \\
B Artifacts & Yes & Listed below the table \\
B1 Citations & Yes & Sections 2 and 3.1, Appendices C, H \\
B2 Licences & Yes & Ethics statement, Appendix H \\
B3 Intended use & Yes & Appendix H \\
B4 Personal information, offensive content & Partly & Ethics statement, Appendix H \\
B5 Documentation & Yes for language, site and register & Sections 3.1 and 3.2, Appendix B \\
B6 Statistics and splits & Yes & Sections 3.1, 3.2 and 3.4, Appendices B, E \\
C1 Parameters, compute & Yes & Sections 3.2 and 3.4, ledger A.15, Ethics statement, Appendices C, D, H \\
C2 Setup, hyperparameters & Yes & Sections 3.2 and 3.4, ledger A.14, ledger A.15, Appendices C, D \\
C3 Descriptive statistics & Yes & Sections 3.4, 4 and 5, Limitations section \\
C4 Packages & Yes & Appendix H, text below the table \\
D Human participants & No & Ethics statement, Appendix H \\
E1 AI assistants & Yes & Ethics statement, Appendix H, S1 \\\bottomrule
\end{tabularx}\caption{Answers to the ACL Responsible NLP Research Checklist. The last column names the sections of the paper that carry each answer.}\end{table}

\textbf{Artifacts and data.} We use Korean Wikipedia discussion comments and Wikisource works, the Blog Authorship Corpus and the ConvoKit Wikipedia Talk Pages Corpus \citep{danescu-niculescu-mizil-2012-echoes}. The base models are Qwen3-8B and Qwen3-32B \citep{yang-2025-qwen3} and Gemma 3 12B \citep{gemma-2025-gemma3}, and the encoders include LUAR and KLUE RoBERTa-base \citep{park-2021-klue}. Qwen3, LUAR and KoELECTRA-v3 are under Apache-2.0, and Gemma 3 is under the Gemma Terms of Use. KLUE RoBERTa-base and ko-sroberta-multitask state no licence. We create the bullet-to-text benchmark with its index files, prompts and rebuild scripts, the M2 verifier and the per-author adapters, and we release no adapter weights. Texts are not redistributed, and author ids are pseudonymous, not anonymous. We report no screening for offensive content, which is why item B4 is answered partly. We ran no human evaluation or annotation, so items D1--D5 do not apply.

\textbf{Compute, statistics and tools.} All local runs used one desktop machine, and the run logs record about 107 hours of wall-clock time for the reported results. The M2 verifiers took 34 hours, the bullet study 54 and the k curve 19, excluding API calls and baselines. We report paired bootstrap intervals at the 95\% level unless a table or sentence names another level, as for the 97.5\% intervals of the verifier comparisons of Section 3.4 and of the word-end verifier H2 against the best baseline. Ledger A.9 lists the level of every comparison. We use MLX, and chrF \citep{popovic-2015-chrf} with β = 2 from sacreBLEU v2.6.0 \citep{post-2018-call}. LLMs are research instruments here: they write the bullets and the normalisations, serve as baselines and act as judges. An AI coding assistant, Claude Code, wrote and ran the experiment code and drafted text under the author's direction. A number-check script re-reads every quoted number from the result files. Further detail per item is in the checklist section of the ledger.

\end{document}